\documentclass{article}
\usepackage[final]{colm2026_conference}

\usepackage{microtype}
\usepackage{hyperref}
\usepackage{url}
\usepackage{booktabs}
\usepackage{amsmath}
\usepackage{amssymb}
\usepackage{amsthm}
\usepackage{graphicx}
\usepackage{xcolor}
\usepackage{multirow}
\usepackage{enumitem}

\usepackage[T1]{fontenc}
\usepackage{lmodern} % Latin Modern (Computer Modern-like) text fonts
\usepackage{inconsolata} % monospace

\usepackage{lineno}

\definecolor{darkblue}{rgb}{0, 0, 0.5}
\hypersetup{colorlinks=true, citecolor=darkblue, linkcolor=darkblue, urlcolor=darkblue}

\newcommand{\vect}[1]{\boldsymbol{#1}}
\newcommand{\mat}[1]{\mathbf{#1}}
\newcommand{\R}{\mathbb{R}}

\title{Spillover-Aware Multi-Value Steering for Pluralistic LLM Alignment}

\author{Weici Pan, Xander Barron, Jiawei Zhou \& Zhenhua Liu \\
Stony Brook University\\
\texttt{\{weici.pan, xander.barron, jiawei.zhou.1, zhenhua.liu\}@stonybrook.edu} 
}

\begin{document}

\maketitle

\begin{abstract}
Activation steering controls LLM behavior at inference time by adding learned directions to hidden states, but existing methods handle one concept at a time. Pluralistic alignment, where different stakeholders need different value emphases, requires steering multiple dimensions simultaneously. We show that naive steering produces substantial spillover: the effect intended for one value leaks into others. This parallels the treatment-versus-spillover decomposition in causal inference. We trace spillover to geometric entanglement of steering directions, captured by their Gram matrix, and derive a zero-cost correction from an activation-norm-penalized objective that decouples each direction's contribution exactly. Our end-to-end pipeline requires no fine-tuning, no reward model, and no manual prompt engineering: given only domain questions, it automatically discovers value dimensions, extracts directions, diagnoses entanglement, and applies corrected steering. On climate discourse, the correction improves the net steering effect from +5.9\% to +14.0\%, validated over 100,000 pairwise judgments.
\end{abstract}

\section{Introduction}
\label{sec:intro}

When asked questions like ``Should we invest in nuclear energy to combat climate change?'', different stakeholders expect different emphases. An economist wants cost-benefit analysis weighing construction costs against decarbonization timelines. An environmental justice advocate wants discussion of disproportionate siting impacts on marginalized communities. An intergenerational ethicist wants focus on waste storage obligations spanning millennia. Standard LLM alignment produces a single response reflecting averaged human preferences~\citep{casper2023open}, but pluralistic alignment recognizes that no single response serves all stakeholders~\citep{sorensen2024value, feng2024modular}.

Throughout, a value taxonomy has $K$ dimensions, each with a unit steering direction $d_j$ in activation space; $\mat{D} = [d_1, \ldots, d_K]$ collects them, a stakeholder preference is a vector $\vect{w} \in \R^K$ of requested emphases, and $\mat{G} = \mat{D}^\top \mat{D}$ ($G_{ij} = d_i^\top d_j$) is their Gram matrix.

Several families of methods address this challenge. Training-time approaches~\citep{dai2023safe, rame2024rewarded, li-etal-2025-gradient} require retraining when preferences change. Prompt-based methods~\citep{feng2024modular, jiang2025picaco} avoid retraining but lack continuous control: a preference like $\vect{w} = (0.7, 0.3, 0, 0)$ is not naturally expressible as a single instruction. Activation steering~\citep{turner2023activation, rimsky2024steering} adds learned directions to hidden states with continuous strength and no weight updates, but all existing methods operate on a single concept at a time. No method provides multi-value steering with principled control over which dimensions are affected.

When multiple steering directions are applied simultaneously, the effect intended for one value leaks into others, destroying the dimension-specific control that pluralistic alignment requires.
If every steering request produces the same undifferentiated increase in opinionatedness, then the preference vector~$\vect{w}$ is meaningless in practice, and the promise of pluralistic alignment is vacuous.

We formalize this failure mode as the \emph{net steering effect} (NSE):
\begin{align}
    \mathrm{NSE} = \mathrm{TargetEffect} - \mathrm{Spillover}
    \label{eq:nse_intro}
\end{align}
where $\mathrm{TargetEffect}$ is the probability that a steered response is judged to emphasize the target dimension more than an unsteered baseline, and $\mathrm{Spillover}$ is the corresponding probability averaged over non-target dimensions. $\mathrm{NSE} \approx 0$ means steering increases all dimensions equally and provides no specific control; $\mathrm{NSE} > 0$ means the target dimension is preferentially affected, the control that pluralistic alignment requires.

Existing activation steering work reports target effect and never measures non-target dimensions, so spillover has remained invisible. We find that naively combining steering directions, by adding a weighted sum of the $d_j$ to the hidden state, produces near-zero NSE: the best uncorrected variant reaches 5.9\%. Increasing the steering strength does not help: it amplifies target and leakage proportionally, because their ratio is fixed by the Gram matrix~$\mat{G}$. This parallels the treatment-versus-spillover decomposition in causal inference~\citep{hudgens2008toward}, where ignoring interference structure leads to overestimation of treatment specificity. 
Correcting the steering coefficients with $\mat{G}^{-1}$ ensures that the perturbation projects as exactly~$w_j$ along each direction~$d_j$, for every~$j$ simultaneously.  

Our contributions are as follows.

\begin{enumerate}[leftmargin=*]

    \item We show that existing activation steering methods lack dimension-specific control, which is a prerequisite for pluralistic alignment, and we formalize this gap as the net steering effect (NSE). All tested methods produce near-zero NSE: they increase value emphasis uniformly rather than specifically.

    \item We develop an end-to-end pipeline from domain questions to corrected multi-value steering, requiring no fine-tuning, reward model, or manual prompt engineering. The pipeline automatically discovers value dimensions via probe classifiers, extracts contrast-based steering directions, diagnoses their geometric entanglement via the Gram matrix, and applies the $\mat{G}^{-1}$~corrected intervention using only standard tools.

    \item We exploit the geometric structure among steering directions to restore dimension-specific control, achieving 14.0\% NSE versus 5.9\% for uncorrected methods on a climate discourse testbed with 100{,}000+ pairwise judgments. This enables the core promise of pluralistic alignment: stakeholders with different preference vectors receive measurably different value emphases.

\end{enumerate}

\section{Related work}
\label{sec:related}

\textbf{Pluralistic alignment (training-time).}
Training-time multi-value alignment methods achieve strong results but require retraining whenever stakeholder preferences change.
RLHF~\citep{ouyang2022training} and DPO~\citep{rafailov2023direct} align to a single aggregated preference, implicitly fixing one weighting over values. Multi-objective extensions decompose this into separate objectives: Safe RLHF~\citep{dai2023safe} separates helpfulness from harmlessness, MODPO~\citep{zhou2024modpo} extends to $K$~objectives via margin-based DPO, Rewarded Soups~\citep{rame2024rewarded} interpolates reward-specific weight checkpoints post-hoc, and PAL~\citep{chen2025pal} learns $K$~preference prototypes from heterogeneous annotators for reward modeling. GAPO~\citep{li-etal-2025-gradient} and OrthAlign~\citep{lin2026orthalign} reduce inter-objective interference during training through gradient orthogonalization and orthogonal parameter decomposition, respectively, conceptually related to our $\mat{G}^{-1}$ decorrelation, though operating in gradient or parameter space rather than activation space. Changing~$\vect{w}$ after deployment requires retraining or maintaining $K$~separate checkpoints.

\textbf{Pluralistic alignment (inference-time).}
Inference-time methods avoid retraining but operate in prompt space, where fine-grained continuous control over value emphasis is difficult to express.
Modular Pluralism~\citep{feng2024modular} routes among community-specific LMs via three pluralism modes. PICACO~\citep{jiang2025picaco} optimizes a meta-instruction navigating multiple values via total correlation, handling up to 8~values. ValueFlow~\citep{kim2026valueflow} provides calibrated intensity control using anchor-based ranking evaluation. These methods select or compose text instructions rather than intervening on internal representations, so a preference like $\vect{w} = (0.7, 0.3, 0, 0)$ is not naturally expressible as a single prompt (Appendix~\ref{app:prompting}). Our method occupies a distinct point in this design space: inference-time like these approaches, but operating in activation space with continuous~$\vect{w}$.

\textbf{Activation steering (single-concept).}
Existing activation steering methods handle one concept at a time and never evaluate whether non-target behaviors are inadvertently affected, leaving the spillover problem invisible.
ActAdd~\citep{turner2023activation} established the additive intervention paradigm: add the mean activation difference from a contrast pair to steer behavior. CAA~\citep{rimsky2024steering} robustifies this by averaging over datasets of contrast pairs. RepE~\citep{zou2023representation} uses PCA on contrastive activations, and ITI~\citep{li2023iti} intervenes on probe-identified attention heads for truthfulness. Subsequent work extends the paradigm to conditional activation~\citep{lee2025cast}, input-adaptive scaling~\citep{wang2025sadi, rodriguez2025dsas}, and SAE-based feature steering~\citep{soo2025fgaa}. ConVA~\citep{jin2025internal} trains and gates one value vector at a time, deferring multi-value control. Evaluation throughout measures only whether the target behavior changed; our NSE metric (\S\ref{sec:nse_def}) is designed to surface what it misses.

\textbf{Multi-property steering.}
The closest prior work identifies the multi-property interference problem and addresses it heuristically or through training. \citet{weij2024multi} show that naive vector summation for multiple properties fails due to feature entanglement and recommend injecting different vectors at different layers, a heuristic that avoids interference in some cases but does not scale beyond 2--3~properties and provides no guarantee. Persona Vectors~\citep{chen2025persona} extract per-trait steering vectors and observe inter-trait correlations (cosine similarities up to 0.3--0.5 across Big Five personality dimensions), and NeVA~\citep{yang2026controllable} observes cross-value leakage when editing value-relevant neurons; both treat it as a finding rather than a problem to solve. MSRS~\citep{jiang2025msrs} separates shared and attribute-private subspaces with a trained steering module, and steering tokens~\citep{radevski2026compositional} distill behaviors into learned input embeddings under an orthogonality regularizer; both buy disentanglement with training. Closest in aim, COAST~\citep{nguyen2026minimizing} attenuates the collateral movement of a steering direction by weighting the activation perturbation with the forward second-moment matrix of activations, one direction at a time. Our work formalizes the interference structure via the Gram matrix~$\mat{G}$, uses its condition number $\kappa(\mat{G})$ as a quantitative diagnostic, and derives the $\mat{G}^{-1}$ correction, which decorrelates all~$K$ directions jointly, in closed form, at inference, with no training. The NSE metric we use to evaluate the correction is inspired by the treatment-versus-spillover decomposition in causal inference~\citep{hudgens2008toward}.
\section{Preliminaries}
\label{sec:prelim}

\subsection{Multi-value activation steering}
\label{sec:steering_setup}

Multi-value steering requires a mapping from a stakeholder's preference to steering coefficients $\vect{\alpha}$. Activation steering modifies an LLM's behavior at inference time by adding learned direction vectors to hidden states. For a single concept, the standard intervention at layer~$\ell$ replaces the hidden state~$h$ with $h' = h + \alpha d$, where $d$~is a learned direction and $\alpha$~is a scalar strength~\citep{turner2023activation, rimsky2024steering}. We adopt the additive form following this established paradigm; alternative interventions such as multiplicative scaling or subspace projection exist~\citep{postmus2024conceptors} but have shown inferior steering effectiveness in our setting (\S\ref{sec:experiments}, Appendix~\ref{app:dead}).

To steer $K$~value dimensions simultaneously, we generalize to:
\begin{align}
    h' = h + \sum_{k=1}^{K} \alpha_k d_k = h + \mat{D}\vect{\alpha}
    \label{eq:intervention}
\end{align}
where $\mat{D} = [d_1, \ldots, d_K] \in \R^{n \times K}$ collects the $K$~steering directions and $\vect{\alpha} \in \R^K$ are the steering coefficients.

A stakeholder expresses a preference vector $\vect{w} \in \R^K$ over value dimensions. For example, $\vect{w} = e_3$ means ``emphasize only the third value dimension,'' while $\vect{w} = (0.5, 0, 0.5, 0)$ requests equal emphasis on the first and third. We seek a mapping from preferences to steering coefficients and parameterize this mapping as linear:
\begin{align}
    \vect{\alpha}^* = \mat{H}\vect{w}
    \label{eq:Hw}
\end{align}
where $\mat{H} \in \R^{K \times K}$ is a \emph{correction matrix}. We restrict~$\mat{H}$ to be linear: $K^2 = 16$ parameters are estimable from limited data, and the exact decorrelation property (\S\ref{sec:ginv}) holds precisely under linear~$\mat{H}$.
All existing single-concept steering methods implicitly set $\mat{H} = \mat{I}$, equating the preference vector directly with the steering coefficients.

\subsection{The Gram matrix and geometric leakage}
\label{sec:gram_def}

The Gram matrix~$\mat{G}$ quantifies how much energy naive steering leaks into non-target dimensions. For $K$~steering directions $d_1, \ldots, d_K$, $\mat{G} \in \R^{K \times K}$ captures their pairwise geometry. We define~$\mat{G}$ from the directions as actually used in the intervention:
\begin{align}
    G_{ij} = d_i^\top d_j
    \label{eq:gram}
\end{align}
When all directions are unit-normalized (as in our experiments; see \S\ref{sec:setup} for details), $\mat{G}$~reduces to the cosine similarity matrix and its diagonal entries are~1. The decorrelation property derived in \S\ref{sec:ginv} holds for $\mat{G} = \mat{D}^\top \mat{D}$ regardless of normalization convention, as long as the same~$\mat{D}$ is used in both the intervention~\eqref{eq:intervention} and the Gram matrix computation. We report condition numbers $\kappa(\mat{G})$ using the cosine form for interpretability: $\kappa = 1$ means perfectly orthogonal directions, and larger values indicate greater entanglement.

Consider steering with $\mat{H} = \mat{I}$ (the existing naive baseline). The projection of the perturbation $\mat{D}\vect{w}$ onto direction~$d_j$ is:
\begin{align}
    d_j^\top (\mat{D}\vect{w})
    = (\mat{D}^\top \mat{D} \, \vect{w})_j
    = (\mat{G}\vect{w})_j
    = \sum_k G_{jk} w_k
    \label{eq:leakage}
\end{align}
When a stakeholder requests emphasis only on dimension~$i$ (i.e., $\vect{w} = e_i$), the projection onto a non-target direction $j \neq i$ equals~$G_{ji}$, not zero. This is \emph{geometric leakage}: the intended single-dimension intervention activates other dimensions in proportion to their inner products with the target. For instance, if $G_{ji} = 0.7$, then 70\% of the target activation leaks into dimension~$j$. The condition number $\kappa(\mat{G})$ summarizes the overall severity of this problem across all dimensions and all possible preference vectors.

\subsection{Measuring steering quality: net steering effect}
\label{sec:nse_def}

The Gram matrix predicts geometric leakage, but verifying its consequences requires measuring the actual effect on generated text. We define the net steering effect (NSE), which separates a method's targeted impact from its unintended spillover. We compare steered responses against unsteered baselines using \textbf{pairwise preference judgments}: given the same question, a judge determines which of the two responses places more emphasis on a specified value dimension. We adopt pairwise comparison because absolute scoring lacks the resolution to detect subtle emphasis shifts ($R^2 < 0.03$ on the same data; see Appendix~\ref{app:eval}).

For a method that steers toward target dimension~$i$, we define the target effect to be the probability that the steered response is judged to emphasize the target dimension more than the unsteered baseline:
\begin{align}
    \mathrm{TargetEffect}_i = P(\text{steered} \succ \text{base} \mid \text{judged on dim } i)
    \label{eq:target}
\end{align}
A target effect above 50\% indicates that steering successfully increases emphasis on the intended dimension. %Ties are counted as 0.5 for both sides.

We define spillover to be the average probability that the steered response is judged to emphasize non-target dimensions more than the baseline:
\begin{align}
    \mathrm{Spillover}_i = \frac{1}{K-1} \sum_{j \neq i} P(\text{steered} \succ \text{base} \mid \text{judged on dim } j)
    \label{eq:spillover}
\end{align}
Spillover above 50\% indicates that steering toward dimension~$i$ unintentionally increases emphasis on other dimensions, precisely predicted by the off-diagonal entries of~$\mat{G}$.

The net steering effect (NSE) is the difference between target effect and spillover:
\begin{align}
    \mathrm{NSE}_i = \mathrm{TargetEffect}_i - \mathrm{Spillover}_i
    \label{eq:nse}
\end{align}
We report the average across all $K$~target dimensions: $\mathrm{NSE} = \frac{1}{K} \sum_i \mathrm{NSE}_i$. 
An ideal method has high target effect and spillover at or below 50\%, yielding large positive NSE. Conversely, a method with target effect equal to spillover ($\mathrm{NSE} = 0$) provides no dimension-specific control: it increases emphasis uniformly across all dimensions, regardless of the stakeholder's intent. Single-concept steering papers typically report only target effect (effectiveness); we argue that NSE is the appropriate metric for multi-value settings, where practical utility depends on specificity and not on effectiveness alone.

\section{Method}
\label{sec:method}

Direction extraction methods differ widely in how entangled their directions are, and the more entangled the directions, the more energy naive steering leaks into non-target dimensions. The Gram matrix condition number $\kappa(\mat{G})$ measures this before any generation is run.

%\textbf{Direction extraction methods.}
We extract four families of steering directions from the same model (Llama-3.1-8B-Instruct) and data (1,255 climate questions with 4-dimensional value labels), each capturing a different geometric object in activation space:
\begin{itemize}[leftmargin=*,itemsep=2pt]
    \item \emph{Probe} (logistic regression): the weight vector of a binary classifier trained to detect whether a question involves value~$k$. This is a \emph{discriminative} direction. It finds the optimal linear boundary separating ``involves $k$'' from ``does not involve $k$.''
    \item \emph{Contrast} (CAA-style): the mean activation difference between value-specific persona-prompted and neutral-prompted forward passes, averaged over all 1,255 questions~\citep{rimsky2024steering}. This is a \emph{generative} direction. It captures how the model's internal state shifts when generating value-emphasizing content. Persona Vectors~\citep{chen2025persona} use an equivalent extraction procedure.
    \item \emph{ActAdd}: a single contrast pair rather than a dataset average~\citep{turner2023activation}.
    \item \emph{RepE}: PCA on contrastive activations, taking PC1~\citep{zou2023representation}.
\end{itemize}

Computing the Gram matrix $\mat{G}$ for each family reveals a dichotomy (Appendix~\ref{app:gram}). Generative directions (ActAdd, Contrast, RepE) are highly entangled: $\kappa(\mat{G})$ ranges from 11.3 to 16.4, with off-diagonal entries of 0.70 to 0.85. Discriminative directions (Probe) are near-orthogonal: $\kappa(\mat{G}) = 1.75$, with off-diagonal entries below 0.20.

The two camps also lie in near-orthogonal subspaces relative to each other (probe--contrast cosine 0.01--0.09; Appendix~\ref{app:detection_control}): the directions that detect a value are not the directions that steer it.
$\mat{G}$ therefore does two things. It predicts how much spillover a given set of directions will produce, and it supplies the correction.

\subsection{Geometric correction}
\label{sec:ginv}

Setting $\mat{H} = \mat{G}^{-1}$ makes each direction's contribution exactly equal to the stakeholder's weight, eliminating all cross-dimensional leakage in the direction geometry. 
For $\vect{\alpha} = \mat{G}^{-1}\vect{w}$, the projection of the perturbation onto direction~$d_j$ satisfies:
\begin{align}
    d_j^\top (\mat{D} \, \mat{G}^{-1} \vect{w})
    = (\mat{D}^\top \mat{D} \, \mat{G}^{-1} \vect{w})_j
    = (\mat{G} \, \mat{G}^{-1} \vect{w})_j
    = w_j
    \label{eq:decorrelation}
\end{align}
Compare with naive steering ($\mat{H} = \mat{I}$), where the projection is $(\mat{G}\vect{w})_j = \sum_k G_{jk} w_k \neq w_j$ (Eq.~\ref{eq:leakage}). The property is exact: it holds for any~$\vect{w}$, any steering strength, and does not depend on linearization, small-$\vect{\alpha}$ assumptions, or any property of the downstream generation process. The intuition is that $\mat{G}^{-1}$ constructs the dual (reciprocal) basis of the direction set: each dual vector is orthogonal to all original directions except its paired one, so their contributions decouple. The same correction falls out of a norm-regularized alignment objective (Appendix~\ref{app:discussion_full}). The correction is demand-driven: by Eq.~\ref{eq:decorrelation} it drives cross-dimensional movement to zero only on dimensions where $w_j = 0$, and delivers exactly the requested emphasis wherever $w_j > 0$. Two genuinely related values can therefore be moved together by giving both non-zero weight; tolerance for spillover is expressed through the shape of~$\vect{w}$, with no separate knob (Appendix~\ref{app:mixed_w}). An adaptive relaxation $\mat{H}_\beta = (\mat{G} + \beta\mat{I})^{-1}$, for preferences whose active directions are themselves strongly collinear, is developed in Appendix~\ref{app:hbeta}. A theoretically stronger correction exists that additionally uses an estimate $\mat{M}$ of how judged emphasis responds causally to steering. At the perturbation scales where steering actually works, that estimate is noise-dominated (per-question coefficient of variation $11.8$) and the Fisher-weighted Gram matrix it induces collapses to order $10^{-4}$, so the resulting coefficients are unusable. The obstruction is a property of the additive intervention regime rather than of the objective, and $\mat{G}^{-1}$ is the member of this family that survives it (Appendix~\ref{app:full_H}).

At generation time, the only change is replacing $\vect{w}$ with $\mat{G}^{-1}\vect{w}$ before computing the steering vector $\mat{D}\vect{\alpha}$. No additional training, reward model, generation, or evaluation is needed. 
\subsection{Pipeline}
\label{sec:pipeline}

Our end-to-end pipeline takes a set of domain questions and produces corrected multi-value steering without fine-tuning, reward models, or manual prompt engineering. The pipeline uses a dual-track design: probe classifiers discover and validate the value taxonomy, while contrast-based directions provide the actual steering vectors. This division reflects the detection--control gap established in \S\ref{sec:method}: probe directions are geometrically clean but causally inert; contrast directions are steerable but entangled, and the $\mat{G}^{-1}$~correction resolves their entanglement exactly. %Figure~\ref{fig:pipeline} illustrates the four stages.

\textbf{Stage 1: Discover value dimensions.}
An LLM annotates domain questions on candidate value dimensions seeded from moral psychology literature~\citep{schwartz2012overview, haidt2004intuitive}. Correlation-based compression reduces 12~candidates to 4~core dimensions (Accuracy, TechEcon, SocialJustice, FutureEthics) capturing 82.5\% of variance. Unsupervised clustering does not recover value structure (Appendix~\ref{app:clustering}); full construction details are in Appendix~\ref{app:taxonomy}.

\textbf{Stage 2: Extract directions.}
Logistic regression probes on last-token activations validate the taxonomy (mean F1~=~0.77; cross-domain transfer F1~=~0.74; Appendix~\ref{app:probes}). Contrast directions, the mean activation difference between persona-prompted and neutral-prompted passes~\citep{rimsky2024steering}, form the steering matrix~$\mat{D}$. Probes validate; contrast directions steer (\S\ref{sec:method}).

\textbf{Stage 3: Diagnose and correct.}
Compute $\mat{G} = \mat{D}^\top \mat{D}$ from the contrast directions and inspect $\kappa(\mat{G})$. If $\kappa$ is large, naive steering will suffer from spillover (\S\ref{sec:method}). Apply $\mat{H} = \mat{G}^{-1}$ (\S\ref{sec:ginv}). 

\textbf{Stage 4: Steer.}
For any stakeholder preference~$\vect{w}$, compute $\vect{\alpha} = \mat{G}^{-1}\vect{w}$, form the perturbation $\mat{D}\vect{\alpha}$, and add it to hidden states during generation. 

\section{Experiments}
\label{sec:experiments}

\subsection{Setup}
\label{sec:setup}

We evaluate on two scales of the Llama family to test both the correction's effectiveness and its consistency across model capacity. The primary experiments use Llama-3.2-3B-Instruct (layer~14, $\alpha = 26$); cross-scale validation uses Llama-3.1-8B-Instruct (layer~20, $\alpha = 20$). Layers are selected as those maximizing mean probe F1 across dimensions. The correction's advantage is robust to this choice: it improves NSE at layers 10, 14, and 18 alike (Appendix~\ref{app:layer_sweep}). Steering strength is calibrated so that $\|\mat{D}\vect{\alpha}\| / \|h\| \approx 15\%$, within the effective range identified in prior work~\citep{turner2023activation, rimsky2024steering}.
Test sets consist of 200~questions (3B) and 100~questions (8B) drawn from the 1,255 StackExchange climate questions labeled on four value dimensions (\S\ref{sec:pipeline}; Appendix~\ref{app:taxonomy}). Cross-architecture validation on Qwen2.5-7B-Instruct is in Appendix~\ref{app:qwen}.

We evaluate the four direction extraction methods described before. Each family produces its own Gram matrix~$\mat{G}$ and corresponding correction~$\mat{G}^{-1}$. Combining four direction types with two correction strategies ($\mat{H} = \mat{I}$ and $\mat{H} = \mat{G}^{-1}$) yields eight methods. A ninth baseline, layer separation, assigns each value to a different layer following \citet{weij2024multi}. All directions are unit-normalized. 

For each method and target dimension, we generate a steered response and compare it against an unsteered baseline (same question, identical generation parameters). Generation uses greedy decoding with a maximum of 256~tokens. 
The judge model is DeepSeek-Chat; three judgments per comparison with randomized A/B position and majority vote, with ties counting as~0.5. All experiments steer with one-hot preference vectors $\vect{w} = e_i$, the most demanding test of dimension-specific control: each request targets exactly one dimension and should leave the other three unaffected. Generalization to mixed~$\vect{w}$ follows from linearity of the intervention (Eq.~\ref{eq:intervention}) and is verified empirically in Appendix~\ref{app:mixed_w}. Sensitivity to~$\alpha$ is analyzed in Appendix~\ref{app:alpha_sweep}. Full configuration counts are in Appendix~\ref{app:config}.

\subsection{Main results}
\label{sec:main_results}

Table~\ref{tab:main_3b} reports the primary results on Llama-3.2-3B-Instruct. The corrected method contrast\_ginv achieves NSE~=~14.0\%, compared to 5.9\% for its uncorrected counterpart ($z = 3.40$, $p < 0.001$), and no uncorrected method exceeds 6\%.

\begin{table}[t]
\begin{center}
\small
\begin{tabular}{llcrrr}
\toprule
Direction & $\kappa(\mat{G})$ & $\mat{H}$ & Target & Spillover & NSE \\
\midrule
\multirow{2}{*}{Contrast} & \multirow{2}{*}{24.4} & $\mat{G}^{-1}$ & 56.8\% & 42.9\% & \textbf{+14.0\%} \\
 & & $\mat{I}$ & 55.2\% & 49.4\% & +5.9\% \\
\midrule
\multirow{2}{*}{RepE} & \multirow{2}{*}{7.8} & $\mat{G}^{-1}$ & 57.5\% & 49.5\% & \textbf{+7.8\%} \\
 & & $\mat{I}$ & 53.3\% & 49.6\% & +3.7\% \\
\midrule
\multirow{2}{*}{ActAdd} & \multirow{2}{*}{14.2} & $\mat{G}^{-1}$ & 53.4\% & 46.9\% & +6.4\% \\
 & & $\mat{I}$ & 52.1\% & 48.2\% & +3.2\% \\
\midrule
Probe & 1.9 & $\mat{I}$ & 55.5\% & 49.7\% & +5.8\% \\
Layer-sep & --- & heuristic & 50.6\% & 49.1\% & +1.5\% \\
\bottomrule
\end{tabular}
\end{center}
\caption{Net steering effect on Llama-3.2-3B-Instruct (200~questions, 4~target dimensions, 3~judges per comparison). Each generative direction type is shown with and without $\mat{G}^{-1}$ correction. The condition number $\kappa(\mat{G})$ measures direction entanglement (\S\ref{sec:method}).}
\label{tab:main_3b}
\end{table}

The correction improves every generative direction type, confirming that it operates on geometric structure rather than direction-specific properties. Contrast directions, the most entangled, gain $+8.1$~percentage points of NSE from the correction; RepE gains $+4.1$~pp and ActAdd $+3.2$~pp. The layer separation heuristic of \citet{weij2024multi} achieves only 1.5\%, confirming that distributing directions across layers does not address the underlying geometric coupling (\S\ref{sec:method}). Decorrelation leaves judged fluency essentially unchanged and adds only a small perplexity increase over naive steering ($+0.19$ in the median; Appendix~\ref{app:quality}).

The two best corrected methods achieve specificity through complementary mechanisms. Contrast\_ginv suppresses spillover below the 50\% chance level (42.9\%). The $\mat{G}^{-1}$ correction removes the shared component that causes all values to rise together: contrast directions concentrate 70.8\% of variance in a single shared eigenvector (\S\ref{sec:method}), and inverting $\mat{G}$ subtracts exactly this component. RepE\_ginv instead leaves spillover at chance (49.5\%) while the target rises, a pure specificity boost. RepE applies PCA during extraction, which partially removes shared variance before $\mat{G}^{-1}$ is applied, leaving spillover near chance rather than suppressing it further.

The correction's advantage holds along two independent axes. As more values are steered jointly, the Gram matrix becomes more ill-conditioned ($\kappa(\mat{G})$ rises from $24.4$ at $K=4$ to $329.7$ at $K=12$), and naive steering stops working: its macro NSE falls below chance at $K=8$ ($-3.5\%$) and $K=10$ ($-3.4\%$), where the non-target dimensions rise more than the one the stakeholder asked for. The corrected method stays positive at every $K$ ($+9.0\%$, $+6.7\%$, and $+17.8\%$ at $K=8$, $10$, and $12$), with a gain over naive at least as large as at $K=4$ (Appendix~\ref{app:kscaling}). The multi-value regime the correction was built for is the regime in which the uncorrected baseline fails outright. On a from-scratch Medical Ethics domain, whose four principles are near-orthogonal ($\kappa(\mat{G})=5.30$), the correction yields a $+11.7$pp gain (Appendix~\ref{app:medical}). The method helps whether the geometry is adversarial or benign.

\begin{table}[t]
\begin{center}
\small
\begin{tabular}{llrrr}
\toprule
Method & Target dim & Target & Spillover & NSE \\
\midrule
\multirow{4}{*}{contrast\_ginv} & Accuracy & 65.0\% & 48.8\% & +16.3\% \\
 & TechEcon & 53.0\% & 39.6\% & +13.4\% \\
 & SocialJustice & 60.5\% & 37.2\% & +23.2\% \\
 & FutureEthics & 48.8\% & 46.0\% & +2.7\% \\
\midrule
\multirow{4}{*}{repe\_ginv} & Accuracy & 61.8\% & 49.6\% & +12.2\% \\
 & TechEcon & 55.5\% & 41.4\% & +14.1\% \\
 & SocialJustice & 49.5\% & 54.9\% & $-5.4$\% \\
 & FutureEthics & 63.2\% & 52.0\% & +11.2\% \\
\bottomrule
\end{tabular}
\end{center}
\caption{Per-dimension breakdown for the two best corrected methods (Llama-3.2-3B-Instruct, 200~questions). The two methods have complementary strengths across value dimensions. Full breakdown for all 9~methods is in Appendix~\ref{app:perdim}.}
\label{tab:perdim_3b}
\end{table}

Table~\ref{tab:perdim_3b} decomposes the two best methods by target dimension, and the profiles are complementary. Contrast\_ginv achieves its strongest specificity on SocialJustice (NSE~=~+23.2\%) and weakest on FutureEthics (+2.7\%). RepE\_ginv shows the opposite pattern: strongest on FutureEthics (+11.2\%) and TechEcon (+14.1\%), but negative on SocialJustice ($-5.4$\%). This reflects which dimensions each extraction method captures most cleanly: contrast directions, built from persona-prompted generation, carry the SocialJustice signal strongly because the SJ persona produces the most distinctive text; RepE captures the variance axes that align with TechEcon and FutureEthics. Direction type can therefore be selected to match the target dimension. The negative entry also shows that $\mat{G}^{-1}$ is not a universal fix: when the underlying direction does not carry a dimension's signal, geometric correction cannot create it.

Spillover is visible in the generations, and so is its removal. Three pairs from the primary experiment illustrate it; each pairs the naive and the corrected response to the same question, with emphasis rated 0--5 by a separate DeepSeek pass on the target and one off-target dimension (naive then corrected). Asked how to get a landlord to install better insulation, under Accuracy steering, the naive response argues economics (``cost savings, reduced energy consumption, and increased property value''), while the corrected response turns to verifiable inspection detail (``Check the attic, walls, and floors'', ``Take photos and videos as evidence'', ``Measure the R-value''): target Accuracy 3 to 5, off-target TechEcon 3 to 0. Asked how a privatized society would handle disputes over air and water, under SocialJustice steering, the naive response stays in market and compliance terms (air ``as a commodity that can be bought and sold'', disputes with ``regulatory agencies''), while the corrected response centers the people affected (``low-income communities and communities of color without access to clean air and water''): SocialJustice 3 to 5, TechEcon 1 to 0. Asked how Los Angeles pollution compares to the 1970s, under the same steering, the naive response is a purely technical comparison (PM-10 concentrations, California Air Resources Board data), while the corrected response keeps the quantitative record (unhealthy days down 45\% from the 1970s) and reframes it around ``vulnerable communities, especially low-income and minority residents'': SocialJustice 0 to 4 with Accuracy unchanged at 4. Geometric leakage appears as off-target framing in otherwise fluent text; the correction removes the framing while the factual content stays intact.

Instructing the model to emphasize the target value is the zero-training alternative, and a strong one: on the same 200 questions it reaches $+16.9\%$ macro NSE against contrast\_ginv's $+14.0\%$, a difference that is not statistically significant (paired bootstrap $p = 0.23$). The two compose rather than compete: prompting strength varies by a factor of two across natural phrasings and carries no explicit weights, while stacking the instruction with corrected steering at a re-calibrated strength reaches $+27.7\%$ macro NSE at the response quality of corrected steering alone. Appendix~\ref{app:prompting} reports the full comparison, the phrasing sensitivity, and the operating-point ladder.

Because NSE relies on an LLM judge, we validate it across model families. Re-judging the full method matrix with GPT-4o reproduces the method-level results closely (contrast\_ginv $+15.25\%$ against $+15.42\%$), and GPT-5-mini gives a consistent gain on the headline comparison. At the level of individual comparisons the judges agree only moderately (Cohen's $\kappa = 0.393$; ordinal Krippendorff's $\alpha = 0.349$), which reflects the genuine subjectivity of single-item value judgments; the method-level conclusions rest on aggregates over hundreds of comparisons, where the judges agree closely. Appendix~\ref{app:crossjudge} reports the full cross-judge table and ordinal-weighted agreement, and Appendix~\ref{app:human} the human study.

\subsection{Cross-scale validation}
\label{sec:cross_scale}
 
\begin{figure}[t]
\begin{center}
\includegraphics[width=\linewidth]{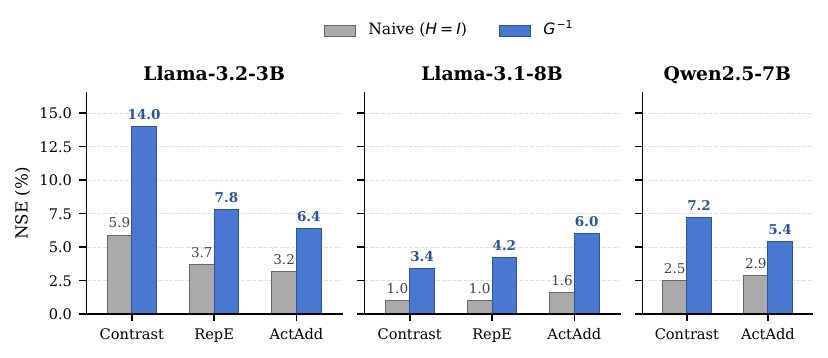}
\end{center}
\caption{$\mat{G}^{-1}$ NSE gain by direction type on three models: Llama-3.2-3B, Llama-3.1-8B, and Qwen2.5-7B (RepE was not extracted for Qwen). Gray bars show naive steering ($\mat{H} = \mat{I}$); colored bars show corrected steering ($\mat{H} = \mat{G}^{-1}$); numbers above bars give NSE in percentage points. The correction improves every direction type on every model. Full 8B and Qwen tables are in Appendices~\ref{app:perdim} and~\ref{app:qwen}.}
\label{fig:crossscale}
\end{figure}
 
Figure~\ref{fig:crossscale} reports the $\mat{G}^{-1}$ NSE gain for each generative direction type on all three models. On Llama-3.1-8B-Instruct (100~questions, layer~20, $\alpha = 20$), the correction improves all three generative direction types: actadd\_ginv achieves NSE~=~6.0\% versus 1.6\% naive ($+4.4$~pp), repe\_ginv 4.2\% versus 1.0\% ($+3.2$~pp), and contrast\_ginv 3.4\% versus 1.0\% ($+2.4$~pp). The qualitative pattern from the 3B experiments replicates: every uncorrected method is near chance, and every corrected method is above it.
 
The best-performing direction type differs across scales: contrast on 3B, ActAdd on 8B. This variation is expected. The direction types differ in extraction procedure (dataset-averaged vs.\ single-pair vs.\ PCA), and which procedure yields the most steerable signal depends on the model's internal geometry, which changes with scale. On 100~questions, the NSE differences among corrected methods (6.0\% vs.\ 4.2\% vs.\ 3.4\%) are within the noise margin, so the ranking itself should be interpreted cautiously. The robust finding is that $\mat{G}^{-1}$ correction is the consistent factor: regardless of which direction type happens to produce the strongest signal, the corrected variant always outperforms its naive counterpart. A practitioner does not need to predict the best direction type in advance; the pipeline extracts all candidate types, applies $\mat{G}^{-1}$ to each, and selects the best by NSE.
 
The pattern extends beyond the Llama family. On Qwen2.5-7B-Instruct, a non-Llama architecture with comparable entanglement ($\kappa_{\mathrm{contrast}} = 11.78$), contrast\_ginv improves NSE by $+4.7$~pp and actadd\_ginv by $+2.5$~pp over their naive counterparts (Figure~\ref{fig:crossscale}, right; Appendix~\ref{app:qwen}).
Steering strength sensitivity is analyzed in Appendix~\ref{app:alpha_sweep}: $\mat{G}^{-1}$ improves NSE across a factor of~3 in~$\alpha$ with no catastrophic collapse.
\subsection{Discussion}
\label{sec:discussion_short}

The decorrelation property (Eq.~\ref{eq:decorrelation}) is exact in direction geometry but generation is nonlinear. The correction nonetheless works because geometric leakage is a primary source of cross-dimensional interference: with off-diagonal Gram entries of 0.7--0.85, most perturbation energy leaks into non-target directions before any nonlinear processing occurs. Removing this deterministic, input-independent leakage is a guaranteed improvement in signal-to-noise ratio regardless of downstream nonlinearities. Any residual interference after correction must arise from mechanisms outside the probe subspace.

The most steerable directions are precisely the most entangled, making correction essential rather than avoidable. 
This detection-control gap motivates the dual-track pipeline: probes validate the taxonomy, contrast directions steer, and $\mat{G}^{-1}$ resolves their entanglement.

$\mat{G}$ also serves as a pre-deployment diagnostic: high $\kappa$ signals poor specificity without correction, and specific off-diagonal entries identify which value pairs will interfere most, with the structure stable across layers ($\|\mat{G}_\ell - \mat{G}_{\ell'}\|_F < 0.08$) and models. The correction extends beyond explicitly pluralistic deployments: a practitioner steering a single target can include auxiliary directions for values they wish to protect and apply $\mat{G}^{-1}$, suppressing spillover onto those dimensions. Limitations, connections to training-time methods, and future directions are in Appendix~\ref{app:discussion_full}.

\section{Conclusion}
\label{sec:conclusion}
 
Multi-value activation steering fails without correcting for the geometric entanglement of steering directions, and the inverse Gram matrix provides that correction exactly. We formalized the spillover problem via the net steering effect (NSE), which separates targeted impact from unintended leakage across value dimensions. The Gram matrix~$\mat{G}$ both diagnoses entanglement severity and points to the fix: $\mat{H} = \mat{G}^{-1}$ eliminates cross-dimensional leakage by construction, for any preference vector and any steering strength, without additional training, reward models, or generation cost. 
On climate discourse with Llama-3.2-3B-Instruct, the correction raises NSE from $+5.9\%$ to $+14.0\%$ ($p < 0.001$), validated over 100{,}000 pairwise judgments, with consistent gains across model scales (Llama-3.1-8B) and architectures (Qwen2.5-7B). 

The pipeline is validated on climate discourse and replicated from scratch on Medical Ethics (Appendix~\ref{app:medical}); broader multi-domain validation remains future work. The full optimal correction $\mat{H}^*$ is numerically unstable in the current additive steering regime, and a unified direction type that is simultaneously discriminative, steerable, and geometrically clean remains open. Extended discussion of limitations and future directions is in Appendix~\ref{app:discussion_full}.

\bibliography{references}
\bibliographystyle{colm2026_conference}

\appendix

\section*{Appendix overview}
 
\begin{itemize}[leftmargin=*,itemsep=1pt]
    \item[\ref{app:taxonomy}.] Value taxonomy construction details
    \item[\ref{app:probes}.] Probe validation
    \item[\ref{app:gram}.] Gram matrix details
    \item[\ref{app:detection_control}.] Detection and control occupy different subspaces
    \item[\ref{app:dead}.] Dead-end experiments
    \item[\ref{app:M_estimation}.] Causal Jacobian and Fisher-weighted Gram matrix estimation
    \item[\ref{app:full_H}.] Full $\mat{H}^*$ test
    \item[\ref{app:eval}.] Evaluation methodology
    \item[\ref{app:alpha_sweep}.] Steering strength sensitivity
    \item[\ref{app:perdim}.] Per-dimension breakdown and full 8B results
    \item[\ref{app:clustering}.] Unsupervised clustering does not recover value structure
    \item[\ref{app:discussion_full}.] Extended discussion
    \item[\ref{app:config}.] Configuration coverage
    \item[\ref{app:mixed_w}.] General preference vectors
    \item[\ref{app:qwen}.] Cross-architecture validation: Qwen2.5-7B-Instruct
    \item[\ref{app:kscaling}.] Scaling the number of value dimensions ($K=8, 10, 12$)
    \item[\ref{app:medical}.] From-scratch replication on a new domain: medical ethics
    \item[\ref{app:crossjudge}.] Judge reliability and cross-family agreement
    \item[\ref{app:layer_sweep}.] Robustness across layers
    \item[\ref{app:quality}.] Response quality under steering
    \item[\ref{app:hbeta}.] Adaptive regularization for collinear preferences
    \item[\ref{app:prompting}.] Instruction prompting as baseline and complement
    \item[\ref{app:human}.] Human validation of the pairwise emphasis judgments
\end{itemize}

\section{Value taxonomy construction details}
\label{app:taxonomy}

The 4-dimension value taxonomy used throughout this paper emerges from thematic grouping of 12~candidate dimensions, informed by inter-dimension correlations and alignment with established moral psychology frameworks.

\textbf{Candidate dimensions and annotation.}
We seed annotation with 12~candidate value dimensions drawn from Schwartz's basic values~\citep{schwartz2012overview}, Moral Foundations Theory~\citep{haidt2004intuitive}, and environmental ethics literature. Table~\ref{tab:12dims} lists all 12~dimensions with their prevalence among 1,255 climate questions from StackExchange, annotated via DeepSeek API with binary labels per dimension.

\begin{table}[tb]
\begin{center}
\small
\begin{tabular}{llr}
\toprule
Dimension & Description & Prev.\ \\
\midrule
Accuracy & Factual precision, scientific evidence & 91.4\% \\
Ecology & Ecosystem health, biodiversity & 41.9\% \\
Equity\_Present & Distributive justice, vulnerable communities & 18.3\% \\
Equity\_Future & Intergenerational obligations & 28.9\% \\
Efficiency & Cost-effectiveness, resource optimization & 34.9\% \\
TechOpt & Technological innovation, engineering solutions & 23.7\% \\
Precaution & Risk aversion, precautionary principle & 27.5\% \\
Liberty & Individual autonomy, freedom from regulation & 11.4\% \\
Procedural & Democratic process, stakeholder participation & 16.4\% \\
Resilience & Adaptive capacity, system robustness & 20.0\% \\
Care & Empathy, welfare of affected populations & 16.0\% \\
Tradition & Cultural continuity, established practices & 11.1\% \\
\bottomrule
\end{tabular}
\end{center}
\caption{Twelve candidate value dimensions and their prevalence in the StackExchange climate dataset ($n = 1{,}255$). Prevalence is the fraction of questions labeled as involving each dimension.}
\label{tab:12dims}
\end{table}

\textbf{Compression to four dimensions.}
Inter-dimension correlations reveal two tightly correlated clusters beyond the standalone Accuracy dimension: a present-oriented justice cluster (Equity\_Present, Care, Procedural, Liberty; pairwise $r = 0.46$--$0.64$) and a future-oriented ecological cluster (Equity\_Future, Precaution, Ecology, Resilience; $r = 0.46$--$0.63$). Efficiency and TechOpt form a smaller pair ($r = 0.60$). Accuracy is anti-correlated with all other dimensions ($r = -0.15$ to $-0.50$), consistent with the intuition that purely factual questions are less likely to engage value trade-offs. We group the 12~dimensions into four based on this correlation structure:

\begin{itemize}[leftmargin=*,itemsep=2pt]
    \item \emph{Accuracy} = \{Accuracy\}
    \item \emph{TechEcon} = \{Efficiency, TechOpt\}
    \item \emph{SocialJustice} = \{Equity\_Present, Care, Procedural, Liberty\}
    \item \emph{FutureEthics} = \{Equity\_Future, Precaution, Ecology, Resilience\}
\end{itemize}

A question receives a positive label for a compressed dimension if it is labeled positive on \emph{any} constituent sub-dimension (logical OR). The resulting prevalence is: Accuracy~91.4\%, TechEcon~38.2\%, SocialJustice~27.6\%, FutureEthics~49.3\%. Appendix~\ref{app:kscaling} reports what the correction does on the uncompressed twelve-dimension set.

\textbf{Post-hoc alignment with established frameworks.}
The compressed dimensions align with established value frameworks without being designed to: Accuracy maps to Schwartz's self-direction and achievement axes (valuing competence and factual mastery); SocialJustice to universalism-benevolence (concern for the welfare of all people); FutureEthics to intergenerational justice and the precautionary principle from environmental ethics; and TechEcon to pragmatic rationality and cost-benefit reasoning. This alignment serves as validation that the data-driven grouping recovers recognizable conceptual structure.

\textbf{Within-group label stability.}
Removing any single sub-dimension from a group changes the compressed labels for at most 14.4\% of questions (TechEcon without Efficiency) and typically less than 3.5\% (Table~\ref{tab:ablation}). The SocialJustice group is particularly robust: no single removal changes more than 3.3\% of labels, indicating that its four sub-dimensions contribute complementarily. In TechEcon and FutureEthics, one sub-dimension (Efficiency and Ecology, respectively) contributes the majority of positive labels, but the remaining sub-dimensions are not redundant. They capture questions that the dominant member misses.

\begin{table}[tb]
\begin{center}
\small
\begin{tabular}{llr}
\toprule
Group & Removed sub-dim & Labels changed \\
\midrule
TechEcon & Efficiency & 14.4\% \\
TechEcon & TechOpt & 3.3\% \\
\midrule
SocialJustice & Equity\_Present & 1.9\% \\
SocialJustice & Care & 3.2\% \\
SocialJustice & Procedural & 3.3\% \\
SocialJustice & Liberty & 1.2\% \\
\midrule
FutureEthics & Ecology & 10.0\% \\
FutureEthics & Equity\_Future & 1.7\% \\
FutureEthics & Precaution & 1.0\% \\
FutureEthics & Resilience & 1.8\% \\
\bottomrule
\end{tabular}
\end{center}
\caption{Within-group label stability. Each row shows the fraction of questions whose compressed label changes when one sub-dimension is removed. Accuracy is a single dimension and is not ablated.}
\label{tab:ablation}
\end{table}

\textbf{Question type clusters.}
K-means clustering on the 4-dimensional compressed label vectors ($K = 4$, silhouette~=~0.725) identifies four interpretable question types:
(1)~\emph{PureAccuracy} (43.6\%): high Accuracy, near-zero on all other dimensions: factual questions with minimal value loading;
(2)~\emph{FutureEthics-focused} (17.6\%): high Accuracy and FutureEthics: questions about long-term consequences;
(3)~\emph{PolicyComplex} (21.8\%): high on all four dimensions: complex policy questions engaging multiple value trade-offs;
(4)~\emph{TechEcon-focused} (17.1\%): high Accuracy and TechEcon: technology and economics questions.
These clusters are referenced in Appendix~\ref{app:discussion_full} (Limitation: input-dependent causal effects), where the per-question causal Jacobian~$\mat{M}(x)$ varies systematically across question types.

Full probe validation including per-dimension F1, cross-domain transfer, and cross-model consistency is in Appendix~\ref{app:probes}.

\section{Probe validation}
\label{app:probes}

Logistic regression probes reliably detect value dimensions across domains, models, and extraction methods. The value taxonomy is learnable and the directions are geometrically meaningful, independent of their role in steering.

\textbf{Per-dimension F1 at best layer.}
Table~\ref{tab:probe_f1} reports 5-fold cross-validated F1 on the 1{,}255 StackExchange climate questions for Llama-3.1-8B-Instruct. Layer~20 is selected as the best layer based on mean F1 across dimensions. Accuracy achieves the highest F1 (0.93), likely because technology and economics questions use distinctive vocabulary. SocialJustice is lowest (0.65), consistent with its high co-occurrence with other dimensions in the taxonomy (Appendix~\ref{app:taxonomy}): questions involving social justice typically also engage accuracy and future ethics, making the binary classification boundary less sharp. F1 is stable across layers 12--24 (mean F1 varies by less than 0.02), indicating that value information is distributed across mid-to-late layers rather than localized.

\begin{table}[tb]
\begin{center}
\small
\begin{tabular}{lcccc}
\toprule
 & Accuracy & TechEcon & SocialJustice & FutureEthics \\
\midrule
F1 (mean $\pm$ std) & $0.93 \pm 0.04$ & $0.79 \pm 0.02$ & $0.65 \pm 0.03$ & $0.76 \pm 0.02$ \\
\bottomrule
\end{tabular}
\end{center}
\caption{Probe F1 scores at layer~20 (Llama-3.1-8B-Instruct, 5-fold CV, $n = 1{,}255$).}
\label{tab:probe_f1}
\end{table}

\textbf{Cross-domain transfer.}
Probes trained on StackExchange climate questions transfer to 3{,}235 Yahoo Answers climate questions with a mean F1 drop of only 0.04: Accuracy~0.88, TechEcon~0.71, SocialJustice~0.68, FutureEthics~0.79 (evaluated on Llama-3.2-1B-Instruct, layer~8). The distribution shift is moderate yet the learned directions generalize, indicating that the value structure captured by probes reflects genuine model representations rather than surface-level stylistic features of the training set.

\textbf{Cross-model consistency.}
The Gram matrix condition number $\kappa(\mat{G}_{\mathrm{probe}})$ decreases with model scale: ${\sim}10$ at 1B parameters, ${\sim}5$ at 3B, ${\sim}2$ at 8B. This trend suggests that larger models develop more geometrically disentangled value representations. The trend is consistent across layers and robust to probe hyperparameter choices (regularization strength~$C$, PCA preprocessing dimension).

\textbf{Direction extraction method comparison.}
We compare five methods for extracting discriminative directions from the same labeled data: logistic regression (LogReg), linear discriminant analysis (LDA), mean difference (MeanDiff), PCA on positive examples (PCA-pos), and PCA on activation differences (PCA-diff). LogReg and LDA converge to similar directions (cosine similarity 0.80--0.84), providing mutual validation: two different optimization objectives recover approximately the same separating hyperplane. PCA-based directions are orthogonal to LogReg/LDA by construction (cosine $< 0.01$). We select LogReg for the main paper because it achieves the highest F1, produces the sparsest weight vectors, and is standard practice in mechanistic interpretability.

\section{Gram matrix details}
\label{app:gram}

This appendix reports the full $4 \times 4$ Gram matrices for all direction types across the three models evaluated in this paper: Llama-3.2-3B-Instruct (layer~14), Llama-3.1-8B-Instruct (layer~20), and Qwen2.5-7B-Instruct (layer~16). Each matrix is computed from unit-normalized directions, so entries are cosine similarities and diagonal entries are~1.

\textbf{Cross-model condition number summary.}
Table~\ref{tab:kappa_all} reports $\kappa(\mat{G})$ for every (model, direction type) pair. The two-camp structure identified in \S\ref{sec:method} holds across all three models: generative directions (Contrast, ActAdd, RepE) have $\kappa \geq 7.8$, while discriminative directions (Probe) have $\kappa \leq 1.9$.

\begin{table}[tb]
\begin{center}
\small
\begin{tabular}{lcccc}
\toprule
Model & Probe & ActAdd & Contrast & RepE \\
\midrule
Llama-3.2-3B & 1.90 & 14.15 & 24.37 & 7.77 \\
Llama-3.1-8B & 1.75 & 11.45 & 16.37 & 11.33 \\
Qwen2.5-7B & 1.74 & 10.73 & 11.78 & --- \\
\bottomrule
\end{tabular}
\end{center}
\caption{Gram matrix condition number $\kappa(\mat{G})$ across models and direction types. Generative directions are consistently an order of magnitude more entangled than discriminative ones. RepE was not extracted for Qwen.}
\label{tab:kappa_all}
\end{table}

\textbf{Eigenvalue concentration.}
The first eigenvalue of generative Gram matrices captures 55--77\% of total variance, reflecting a dominant shared component. Table~\ref{tab:eigvals} reports eigenvalue distributions. For all three models, the first eigenvector of $\mat{G}_{\mathrm{contrast}}$ has near-uniform loadings ($[-0.43, -0.52, -0.52, -0.52]$ on 3B; $[-0.45, -0.52, -0.50, -0.52]$ on 8B; $[-0.29, -0.55, -0.56, -0.54]$ on Qwen), confirming that the shared component is the generic ``be opinionated'' mode rather than any specific value direction. Probe Gram matrices have nearly uniform eigenvalue distributions across all models, with no dominant shared component.

\begin{table}[tb]
\begin{center}
\small
\begin{tabular}{llcccc}
\toprule
Model & Direction & $\lambda_1$ & $\lambda_2$ & $\lambda_3$ & $\lambda_4$ \\
\midrule
\multirow{4}{*}{3B} & Probe & 36.3\% & 24.0\% & 20.6\% & 19.1\% \\
 & ActAdd & 63.6\% & 21.0\% & 10.9\% & 4.5\% \\
 & Contrast & 76.7\% & 13.9\% & 6.3\% & 3.1\% \\
 & RepE & 55.3\% & 22.8\% & 14.8\% & 7.1\% \\
\midrule
\multirow{4}{*}{8B} & Probe & 34.6\% & 24.4\% & 21.2\% & 19.8\% \\
 & ActAdd & 60.3\% & 22.4\% & 12.1\% & 5.3\% \\
 & Contrast & 76.0\% & 11.8\% & 7.5\% & 4.6\% \\
 & RepE & 65.6\% & 21.9\% & 6.7\% & 5.8\% \\
\midrule
\multirow{3}{*}{Qwen} & Probe & 35.0\% & 24.4\% & 20.5\% & 20.1\% \\
 & ActAdd & 62.6\% & 21.9\% & 9.7\% & 5.8\% \\
 & Contrast & 64.6\% & 22.0\% & 7.8\% & 5.5\% \\
\bottomrule
\end{tabular}
\end{center}
\caption{Eigenvalue distribution (percentage of total variance) of Gram matrices. Generative directions concentrate 55--77\% of variance in a single eigenvalue; probe directions distribute variance nearly uniformly.}
\label{tab:eigvals}
\end{table}

\textbf{Full Gram matrices.}
Tables~\ref{tab:gram_3b}--\ref{tab:gram_qwen} report the complete $4 \times 4$ Gram matrices. Columns and rows follow the order: Accuracy (Acc), TechEcon (TE), SocialJustice (SJ), FutureEthics (FE).

\begin{table}[tb]
\begin{center}
\small
\begin{tabular}{l|rrrr}
\toprule
\multicolumn{5}{c}{\emph{3B Probe} ($\kappa = 1.90$)} \\
 & Acc & TE & SJ & FE \\
\midrule
Acc & 1.00 & $-0.12$ & $-0.10$ & $-0.05$ \\
TE & $-0.12$ & 1.00 & $+0.18$ & $+0.21$ \\
SJ & $-0.10$ & $+0.18$ & 1.00 & $+0.21$ \\
FE & $-0.05$ & $+0.21$ & $+0.21$ & 1.00 \\
\midrule
\multicolumn{5}{c}{\emph{3B Contrast} ($\kappa = 24.37$)} \\
 & Acc & TE & SJ & FE \\
\midrule
Acc & 1.00 & $+0.62$ & $+0.54$ & $+0.55$ \\
TE & $+0.62$ & 1.00 & $+0.77$ & $+0.76$ \\
SJ & $+0.54$ & $+0.77$ & 1.00 & $+0.87$ \\
FE & $+0.55$ & $+0.76$ & $+0.87$ & 1.00 \\
\midrule
\multicolumn{5}{c}{\emph{3B ActAdd} ($\kappa = 14.15$)} \\
 & Acc & TE & SJ & FE \\
\midrule
Acc & 1.00 & $+0.55$ & $+0.39$ & $+0.34$ \\
TE & $+0.55$ & 1.00 & $+0.53$ & $+0.44$ \\
SJ & $+0.39$ & $+0.53$ & 1.00 & $+0.81$ \\
FE & $+0.34$ & $+0.44$ & $+0.81$ & 1.00 \\
\midrule
\multicolumn{5}{c}{\emph{3B RepE} ($\kappa = 7.77$)} \\
 & Acc & TE & SJ & FE \\
\midrule
Acc & 1.00 & $+0.28$ & $-0.22$ & $+0.16$ \\
TE & $+0.28$ & 1.00 & $-0.40$ & $+0.58$ \\
SJ & $-0.22$ & $-0.40$ & 1.00 & $-0.66$ \\
FE & $+0.16$ & $+0.58$ & $-0.66$ & 1.00 \\
\bottomrule
\end{tabular}
\end{center}
\caption{Full Gram matrices for Llama-3.2-3B-Instruct (layer~14). Entries are cosine similarities between unit-normalized directions.}
\label{tab:gram_3b}
\end{table}

\begin{table}[tb]
\begin{center}
\small
\begin{tabular}{l|rrrr}
\toprule
\multicolumn{5}{c}{\emph{8B Probe} ($\kappa = 1.75$)} \\
 & Acc & TE & SJ & FE \\
\midrule
Acc & 1.00 & $-0.08$ & $-0.11$ & $-0.05$ \\
TE & $-0.08$ & 1.00 & $+0.16$ & $+0.20$ \\
SJ & $-0.11$ & $+0.16$ & 1.00 & $+0.14$ \\
FE & $-0.05$ & $+0.20$ & $+0.14$ & 1.00 \\
\midrule
\multicolumn{5}{c}{\emph{8B Contrast} ($\kappa = 16.37$)} \\
 & Acc & TE & SJ & FE \\
\midrule
Acc & 1.00 & $+0.62$ & $+0.58$ & $+0.60$ \\
TE & $+0.62$ & 1.00 & $+0.70$ & $+0.80$ \\
SJ & $+0.58$ & $+0.70$ & 1.00 & $+0.76$ \\
FE & $+0.60$ & $+0.80$ & $+0.76$ & 1.00 \\
\midrule
\multicolumn{5}{c}{\emph{8B ActAdd} ($\kappa = 11.45$)} \\
 & Acc & TE & SJ & FE \\
\midrule
Acc & 1.00 & $+0.14$ & $+0.31$ & $+0.35$ \\
TE & $+0.14$ & 1.00 & $+0.67$ & $+0.47$ \\
SJ & $+0.31$ & $+0.67$ & 1.00 & $+0.73$ \\
FE & $+0.35$ & $+0.47$ & $+0.73$ & 1.00 \\
\midrule
\multicolumn{5}{c}{\emph{8B RepE} ($\kappa = 11.33$)} \\
 & Acc & TE & SJ & FE \\
\midrule
Acc & 1.00 & $+0.26$ & $+0.26$ & $+0.26$ \\
TE & $+0.26$ & 1.00 & $+0.73$ & $+0.77$ \\
SJ & $+0.26$ & $+0.73$ & 1.00 & $+0.75$ \\
FE & $+0.26$ & $+0.77$ & $+0.75$ & 1.00 \\
\bottomrule
\end{tabular}
\end{center}
\caption{Full Gram matrices for Llama-3.1-8B-Instruct (layer~20).}
\label{tab:gram_8b}
\end{table}

\begin{table}[tb]
\begin{center}
\small
\begin{tabular}{l|rrrr}
\toprule
\multicolumn{5}{c}{\emph{Qwen Probe} ($\kappa = 1.74$)} \\
 & Acc & TE & SJ & FE \\
\midrule
Acc & 1.00 & $-0.05$ & $-0.10$ & $-0.05$ \\
TE & $-0.05$ & 1.00 & $+0.19$ & $+0.18$ \\
SJ & $-0.10$ & $+0.19$ & 1.00 & $+0.18$ \\
FE & $-0.05$ & $+0.18$ & $+0.18$ & 1.00 \\
\midrule
\multicolumn{5}{c}{\emph{Qwen Contrast} ($\kappa = 11.78$)} \\
 & Acc & TE & SJ & FE \\
\midrule
Acc & 1.00 & $+0.36$ & $+0.24$ & $+0.22$ \\
TE & $+0.36$ & 1.00 & $+0.73$ & $+0.67$ \\
SJ & $+0.24$ & $+0.73$ & 1.00 & $+0.76$ \\
FE & $+0.22$ & $+0.67$ & $+0.76$ & 1.00 \\
\midrule
\multicolumn{5}{c}{\emph{Qwen ActAdd} ($\kappa = 10.73$)} \\
 & Acc & TE & SJ & FE \\
\midrule
Acc & 1.00 & $+0.35$ & $+0.23$ & $+0.23$ \\
TE & $+0.35$ & 1.00 & $+0.70$ & $+0.60$ \\
SJ & $+0.23$ & $+0.70$ & 1.00 & $+0.74$ \\
FE & $+0.23$ & $+0.60$ & $+0.74$ & 1.00 \\
\bottomrule
\end{tabular}
\end{center}
\caption{Full Gram matrices for Qwen2.5-7B-Instruct (layer~16). RepE directions were not extracted for Qwen.}
\label{tab:gram_qwen}
\end{table}

\textbf{Cross-model patterns.}
Three patterns are consistent across all models. First, Accuracy is the least entangled generative direction in every model: its off-diagonal entries are 10--30 percentage points lower than among the other three dimensions. This is consistent with the taxonomy (Appendix~\ref{app:taxonomy}): Accuracy is anti-correlated with all other value dimensions, so the persona shift toward Accuracy shares less with the generic ``opinionated'' mode. Second, the SJ--FE pair has the highest off-diagonal entry in every generative Gram matrix (0.73--0.87), reflecting the thematic overlap between environmental justice and intergenerational ethics in climate discourse. Third, probe Gram matrices are remarkably similar across all three models ($\kappa \in [1.74, 1.90]$, Accuracy anti-correlated with others, TE/SJ/FE weakly positively correlated), suggesting that the discriminative geometry of value representations is an invariant of the value structure rather than a model-specific artifact.

\section{Detection and control occupy different subspaces}
\label{app:detection_control}

Probe directions that detect value emphasis with mean F1~=~0.78 share near-zero cosine similarity with contrast directions that produce measurable steering effects.

Table~\ref{tab:cosine_probe_gen} reports the cosine similarity between each probe direction and its matched generative direction (Contrast, ActAdd, RepE) at layer~20 of Llama-3.1-8B-Instruct. All 12~pairwise cosines fall between 0.01 and 0.09. For comparison, the expected absolute cosine between two random unit vectors in~$\R^{4096}$ is $0.012 \pm 0.009$ (empirical, $10^5$~samples), with 95th percentile~0.031. The observed probe--contrast cosines are not significantly above this random baseline ($p = 0.89$, one-sided). The full cross-pair matrix $|\cos(\text{probe}_i, \text{contrast}_j)|$ for $i \neq j$ is also uniformly below~0.06, confirming that the near-orthogonality is not an artifact of dimension matching.

\begin{table}[tb]
\begin{center}
\small
\begin{tabular}{lcccc}
\toprule
Dimension & Contrast & ActAdd & RepE & Random \\
\midrule
Accuracy       & 0.093 & 0.082 & 0.082 & \\
TechEcon       & 0.019 & 0.024 & 0.061 & \\
SocialJustice  & 0.020 & 0.057 & 0.034 & \\
FutureEthics   & 0.011 & 0.027 & 0.028 & \\
\midrule
Mean           & 0.036 & 0.048 & 0.051 & $0.012 \pm 0.009$ \\
\bottomrule
\end{tabular}
\end{center}
\caption{Cosine similarity between probe and generative directions at layer~20 (Llama-3.1-8B-Instruct). All values are comparable to the random baseline in~$\R^{4096}$, indicating that discriminative and generative directions occupy near-orthogonal subspaces.}
\label{tab:cosine_probe_gen}
\end{table}

This gap is stable across layers. Across layers 8--28, the mean absolute cosine between probe and contrast directions remains in the range 0.025--0.037, never exceeding 0.10 for any individual dimension--layer pair. Meanwhile, $\kappa(\mat{G}_{\mathrm{probe}})$ stays within 1.74--1.88 and $\kappa(\mat{G}_{\mathrm{contrast}})$ ranges from 13 to 24, confirming that the two-camp entanglement structure (\S\ref{sec:method}) is also layer-stable.

Despite their near-orthogonality to probe directions, contrast-based personas causally shift value emphasis with high specificity. Table~\ref{tab:persona_lift} reports the lift matrix from a confound check experiment: for each of the four value personas, we generate responses to 20~climate questions and score them on all four dimensions (0--10 absolute scale, LLM judge). The diagonal entries are uniformly positive (mean lift~=~2.20 points), indicating that each persona increases emphasis on its target dimension. The off-diagonal entries are predominantly negative (mean~=~$-0.32$), indicating mild suppression of non-target dimensions. All four personas produce specific, dimension-targeted effects at the text level. The one notable positive off-diagonal entry (SocialJustice persona lifting FutureEthics by~0.90) reflects thematic overlap between environmental justice and intergenerational ethics in climate discourse.

\begin{table}[tb]
\begin{center}
\small
\begin{tabular}{lcccc}
\toprule
Persona $\downarrow$ \textbackslash\ Judged $\rightarrow$ & Acc & TE & SJ & FE \\
\midrule
Accuracy       & $\mathbf{+0.65}$ & $-0.25$ & $-0.25$ & $+0.10$ \\
TechEcon       & $-1.00$ & $\mathbf{+2.55}$ & $-0.25$ & $-0.50$ \\
SocialJustice  & $-0.75$ & $-0.40$ & $\mathbf{+2.80}$ & $+0.90$ \\
FutureEthics   & $-0.80$ & $-0.65$ & $+0.05$ & $\mathbf{+2.80}$ \\
\bottomrule
\end{tabular}
\end{center}
\caption{Persona generation lift matrix. Entry~$(i,j)$ is the mean score under persona~$i$ minus the neutral baseline score, judged on dimension~$j$. $n = 20$~questions, 0--10 absolute scoring. Diagonal entries (bold) show that each persona specifically increases its target dimension; off-diagonal entries are predominantly negative.}
\label{tab:persona_lift}
\end{table}

This lift matrix measures the effect of text-level persona prompting, not activation steering. It establishes that the generative shift captured by contrast directions is causal and dimension-specific.

The gap between detection and control reflects a fundamental difference in what the two direction types represent geometrically. Probe directions are normal vectors to classification boundaries: they encode \emph{where} value-relevant information resides in activation space, optimized to separate ``involves value~$k$'' from ``does not involve value~$k$.'' Contrast directions capture the mean activation shift during value-emphasizing generation: they encode \emph{how} the model's internal state changes when producing value-laden text. The near-zero cosine means these are not merely different magnitudes of the same direction. A medical analogy clarifies the distinction: a diagnostic classifier might detect tumors from imaging features, but the biological pathways driving tumor growth are entirely different from the features used for detection. Knowing the diagnostic boundary does not reveal the causal mechanism; similarly, knowing where value information is linearly separable does not tell the model how to generate differently.

This gap motivates the dual-track pipeline (\S\ref{sec:pipeline}): probes discover and validate the value taxonomy (high detection accuracy, near-orthogonal geometry), while contrast directions perform the actual steering (high steerability, entangled geometry that $\mat{G}^{-1}$ corrects). A unified direction type that is simultaneously discriminative and steerable would simplify the framework; understanding when and why detection--control alignment fails is an open question with implications beyond multi-value steering.

\section{Dead-end experiments}
\label{app:dead}

Three alternative intervention mechanisms were tested and rejected before arriving at the additive steering framework used in this paper. Each fails for a different reason, and the failures collectively motivate the design choices described in \S\ref{sec:steering_setup}.

Conceptor-based steering~\citep{postmus2024conceptors} fails to produce positive diagonal lifts. It projects activations onto or away from learned subspaces, offering a principled alternative to additive perturbation. We tested three conceptor construction methods (probe-based, persona-based, and difference-based), three aperture settings ($\alpha \in \{1, 4, 16\}$), and five intervention strengths ($\beta \in \{0.1, 0.3, 0.5, 0.7, 0.9\}$) across three layers (16, 20, 24)---211 configurations and 6{,}330 generations total (30~questions each, 12.4~GPU-hours on Llama-3.1-8B-Instruct). The best configuration (method\,=\,diff, aperture\,=\,1, $\beta$\,=\,0.1) achieves a mean diagonal lift of~$-0.17$---that is, steering in the \emph{wrong direction}. For comparison, simple additive probe steering at $\alpha = 5$ achieves $-0.14$, slightly better but also negative. At $\beta \geq 0.5$, conceptor steering crashes generation quality: all four value scores drop to near zero as the model produces repetitive or incoherent output. The root cause is that subspace projection \emph{removes} activation energy rather than redirecting it.

Contrastive decoding destroys generation quality. It upweights tokens that a value-conditioned model prefers over a neutral model, providing value emphasis without modifying activations. We tested four configurations: uniform (equal weight across all values) and per-dimension (Accuracy, TechEcon, SocialJustice). In all cases, the output degenerates into repetitive fragments (e.g., ``TheTheassistantTheTheTheThe\ldots'') with all value scores at zero. The mechanism conflates ``tokens statistically associated with value~$k$'' with ``coherent next tokens'' at any useful contrast strength, coherence collapses. Contrastive decoding has shown success for binary attributes like toxicity reduction~\citep{li2023contrastive}, where the value-relevant token set largely overlaps with the coherent token set; for nuanced multi-value emphasis, this overlap is insufficient.

Layer-specific direction assignment does not improve steering. The heuristic assigning different value directions to different layers recommended by \citet{weij2024multi} to avoid interference between properties does not improve outcomes in our setting. We tested all 24~permutations of assigning 4~value dimensions to 4~candidate layers (8, 12, 20, 24), evaluating each on 20~questions at $\alpha = 3$ with absolute scoring. The score range across all permutations is less than 1.0~point on every dimension (Accuracy: 5.6--6.6, TechEcon: 1.2--2.0, SocialJustice: 0.8--1.4, FutureEthics: 2.0--2.8). No permutation ranking differs significantly from another. The failure is consistent with our Gram matrix analysis (\S\ref{sec:method}): the entanglement structure $\mat{G}$ is stable across adjacent layers ($\|\mat{G}_\ell - \mat{G}_{\ell'}\|_F < 0.08$), so separating directions across layers does not reduce their geometric coupling. In contrast, $\mat{G}^{-1}$~correction addresses entanglement directly in the direction geometry, regardless of layer placement.

\textbf{Single-layer assumption.}
We intervene at a single layer~$\ell$ rather than distributing the intervention across multiple layers. This is a simplifying choice: our ablations (\S\ref{sec:experiments}) show that varying the number of intervention layers (1, 3, 5, or 6~layers) changes steering outcomes by less than 1.0 on a 10-point scale, suggesting that single-layer intervention captures the dominant effect. Multi-layer intervention introduces $L \times K$ coefficients and layer-specific direction matrices, substantially complicating the optimization without proportional benefit in our setting.

\section{Causal Jacobian and Fisher-weighted Gram matrix estimation}
\label{app:M_estimation}

The full optimal correction $\mat{H}^* = \frac{1}{\beta}\tilde{\mat{G}}^{-1}\mat{M}^\top$ (\S\ref{sec:ginv}) requires estimating two quantities: the causal Jacobian~$\mat{M}$ (how steering along direction~$j$ affects value score~$k$) and the Fisher-weighted Gram matrix~$\tilde{\mat{G}}$ (direction inner products weighted by the generation distribution). Both are too small and too noisy to support the full $\mat{H}^*$.

\textbf{Causal Jacobian $\mat{M}$.}
We estimate $M_{kj} = \partial V_k / \partial \alpha_j$ by finite differences: for each direction~$j$, steer with $\alpha_j = \pm\delta$ (holding other coefficients at zero) and measure the value score change via LLM judge scoring (0--10 absolute scale). Table~\ref{tab:M_matrices} reports $\mat{M}$ estimated in both the probe and contrast direction bases. In both cases, the absolute diagonal entries average ${\sim}0.01$. The off-diagonal entries are comparable in magnitude to the diagonal, indicating no clear causal specificity at this resolution.

\begin{table}[tb]
\begin{center}
\small
\begin{tabular}{lrrrr}
\toprule
 & Acc & TE & SJ & FE \\
\midrule
\multicolumn{5}{l}{\emph{$\mat{M}$ in probe basis}} \\
$\alpha_{\text{Acc}}$ & $-0.021$ & $-0.003$ & $-0.007$ & $-0.016$ \\
$\alpha_{\text{TE}}$ & $-0.031$ & $+0.013$ & $+0.013$ & $+0.008$ \\
$\alpha_{\text{SJ}}$ & $-0.003$ & $+0.012$ & $+0.002$ & $+0.014$ \\
$\alpha_{\text{FE}}$ & $-0.005$ & $-0.003$ & $+0.012$ & $+0.016$ \\
\midrule
\multicolumn{5}{l}{\emph{$\mat{M}$ in contrast basis}} \\
$\alpha_{\text{Acc}}$ & $+0.031$ & $+0.014$ & $-0.003$ & $-0.006$ \\
$\alpha_{\text{TE}}$ & $-0.012$ & $-0.005$ & $-0.000$ & $-0.023$ \\
$\alpha_{\text{SJ}}$ & $-0.002$ & $+0.014$ & $-0.016$ & $+0.005$ \\
$\alpha_{\text{FE}}$ & $-0.015$ & $-0.000$ & $+0.003$ & $-0.007$ \\
\bottomrule
\end{tabular}
\end{center}
\caption{Estimated causal Jacobian $\mat{M}$. Entry in row~$j$, column~$k$ is the estimated $\partial V_k / \partial \alpha_j$: how steering along direction~$j$ affects value score~$k$. All entries are $O(10^{-2})$, indicating negligible linear causal effect at the first-order level.}
\label{tab:M_matrices}
\end{table}

\begin{figure}[tb]
\begin{center}
\includegraphics[width=\linewidth]{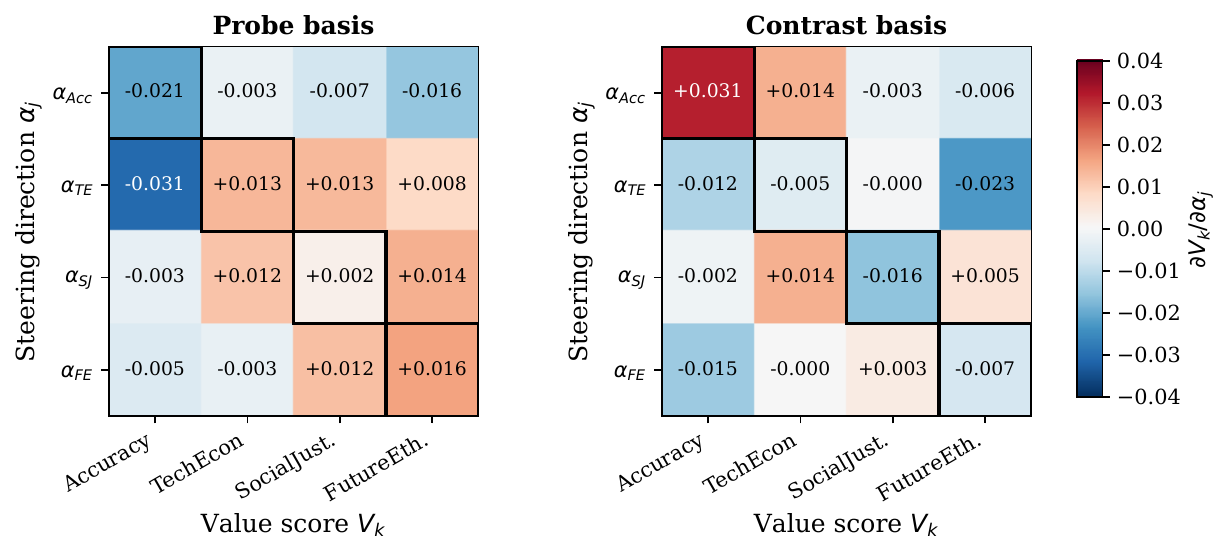}
\end{center}
\caption{Estimated causal Jacobian $\mat{M}$ in probe (left) and contrast (right) direction bases. Black squares mark diagonal entries. A well-behaved $\mat{M}$ would show strong positive diagonal and weak off-diagonal; instead, all entries are $O(10^{-2})$ with no clear diagonal structure.}
\label{fig:M_heatmap}
\end{figure}

Figure~\ref{fig:M_heatmap} visualizes these matrices. Neither basis shows a dominant diagonal: off-diagonal entries are comparable in magnitude to diagonal entries, and several diagonal entries are negative, confirming the absence of linear causal specificity.

\textbf{Per-question variance.}
$\mat{M}$ is not just small. It is wildly inconsistent across questions. Per-question estimates $\mat{M}(x)$ have a coefficient of variation (CV) of~11.8, meaning the standard deviation of the causal effect across questions is an order of magnitude larger than the mean. This high variance has two implications: (i)~a fixed linear $\mat{H}$ cannot capture the input-dependent causal structure, and (ii)~averaging $\mat{M}(x)$ across questions produces a near-zero mean that reflects cancellation rather than absence of effect. The per-question variation is systematic rather than random: it correlates with question type (Appendix~\ref{app:taxonomy}), with PolicyComplex questions showing the largest causal effects and PureAccuracy questions the smallest.

\textbf{Fisher-weighted Gram matrix $\tilde{\mat{G}}$.}
$\tilde{\mat{G}}$ captures direction inner products weighted by the Fisher information of the generation distribution. Its entries are three to four orders of magnitude smaller than the corresponding entries of~$\mat{G}$: $\tilde{G}_{ij} \approx 10^{-4}$ versus $G_{ij} \approx 1$. We estimate $\tilde{\mat{G}}$ via both gradient-based and KL-based approaches; both yield the same order of magnitude, confirming that the smallness is not an artifact of the estimation method. The scale gap between $\mat{G}$ and $\tilde{\mat{G}}$ means that the Fisher weighting is dominated by directions along which the generation distribution is nearly flat.

%\textbf{Why this is a regime problem.}
The smallness of $\mat{M}$ and $\tilde{\mat{G}}$ reflects a fundamental property of additive activation steering, not insufficient data. Meaningful steering requires perturbation magnitude $\|\mat{D}\vect{\alpha}\| / \|h\| \approx 15\%$ ($\alpha = 20$), but the first-order Taylor approximation $V(\vect{\alpha}) \approx V(0) + \mat{M}\vect{\alpha}$ is accurate only where $\|\mat{D}\vect{\alpha}\| / \|h\| < 1\%$. The linear approximation is valid precisely where steering has no effect, and breaks down precisely where steering matters. This gap explains why $\mat{G}^{-1}$ captures the dominant improvement, while the $\mat{M}$-dependent component of the full~$\mat{H}^*$ adds noise rather than signal. A detailed test of the full $\mat{H}^*$ is in Appendix~\ref{app:full_H}.

\section{Full \texorpdfstring{$\mat{H}^*$}{H*} test}
\label{app:full_H}

Computing the full optimal correction $\mat{H}^* = \frac{1}{\beta}\tilde{\mat{G}}^{-1}\mat{M}^\top$ from the estimated $\mat{M}$ and $\tilde{\mat{G}}$ (Appendix~\ref{app:M_estimation}) produces astronomically large steering coefficients that, after rescaling, reduce to a near-uniform direction.

Raw $\mat{H}^*$ produces unusable coefficients. Plugging the estimated $\mat{M}$ and $\tilde{\mat{G}}$ into the closed-form solution yields $\|\vect{\alpha}^*\| \approx 3.4 \times 10^6$. The cause is $\tilde{\mat{G}}^{-1}$: because $\tilde{G}_{ij} \approx 10^{-4}$, its inverse amplifies every entry of $\mat{M}^\top\vect{w}$ by a factor of ${\sim}10^4$, including estimation noise. The regularization parameter $\beta = 68.6$ (estimated from a KL divergence constraint) is far too small to compensate. Rescaled $\mat{H}^*$ is near-uniform. After normalizing $\vect{\alpha}^*$ to match the perturbation norm of $\mat{G}^{-1}$~steering ($\|\mat{D}\vect{\alpha}\| / \|h\| \approx 15\%$), the resulting coefficient vector is $[+1.68, +3.01, +2.10, +2.95]$, near-uniform positive across all four dimensions. This means the rescaled $\mat{H}^*$ effectively says ``increase all values equally,'' which is precisely the undifferentiated opinionatedness that $\mat{G}^{-1}$ is designed to avoid. The $\mat{M}^\top\vect{w}$ component, which should encode value-specific causal information, is dominated by noise; $\tilde{\mat{G}}^{-1}$ amplifies that noise; and rescaling collapses the result to the mean direction.

\textbf{Empirical confirmation.}
We steer 5~test questions with both rescaled $\mat{H}^*$ and $\mat{G}^{-1}$, evaluating via pairwise comparison. $\mat{H}^*$ does not outperform $\mat{G}^{-1}$ on any question. The near-uniform coefficient vector produces the same undifferentiated increase in value emphasis that naive steering ($\mat{H} = \mat{I}$) produces, confirming that the $\mat{M}$-dependent component adds noise rather than signal.

\textbf{Interpretation.}
$\mat{G}^{-1}$ captures the only component of $\mat{H}^*$ that is both computable and useful in the current regime: the geometric correction that decorrelates direction contributions. The remaining component requires a regime where the linear approximation $V(\vect{\alpha}) \approx V(0) + \mat{M}\vect{\alpha}$ is accurate at perturbation magnitudes that actually produce steering effects. As discussed in Appendix~\ref{app:M_estimation}, this regime does not hold for additive activation steering at current perturbation scales. Stronger intervention mechanisms that produce larger, more consistent causal effects per unit perturbation could make $\mat{M}$ and $\tilde{\mat{G}}$ reliably estimable, unlocking the full $\mat{H}^*$.

\section{Evaluation methodology}
\label{app:eval}

This appendix details the pairwise preference judging protocol, judge prompt design, quality checks, and aggregation procedure.

\textbf{Judging protocol.}
For each (method, target dimension, question) triple, we generate a steered response and compare it against an unsteered baseline generated with the same question and identical generation parameters (greedy decoding, 256~max tokens). The comparison is evaluated on all $K = 4$ value dimensions independently: for each dimension, a judge determines which response places more emphasis on that dimension, or whether they are equal. This produces $K$ judgments per comparison. Each judgment is repeated by three judge calls with independently randomized A/B position; the final verdict is the majority vote (ties count as~0.5 for both sides).

\textbf{Judge model and prompt.}
We use DeepSeek-Chat as the judge model (${\sim}\$0.10$ per 1{,}000 calls). The judge receives both responses (labeled A and B, with random order to eliminate position bias) and a natural-language description of the value dimension being evaluated. The prompt asks: ``Which response places more emphasis on [dimension description]? Answer A, B, or tie.'' We provide one-sentence descriptions of each value dimension drawn from the taxonomy (Appendix~\ref{app:taxonomy}), ensuring that the judge evaluates value emphasis rather than response quality.

\textbf{Pairwise versus absolute scoring.}
Multi-value steering effects are subtle shifts in emphasis rather than categorical changes in content, and this drove our protocol choice. On our data, absolute scoring (0--10 per dimension) lacked the resolution to detect them: where pairwise comparison reveals 55--65\% win rates, absolute scoring yields $R^2 < 0.03$ between steering strength and value scores. This matches the comparative-judgment literature: pairwise comparison eliminates the between-judge calibration variance that dominates absolute scales, and it aligns better with human judgment for moderately sized judge models \citep{liusie2024llm}. The choice is nonetheless context-dependent. Pairwise protocols are more vulnerable to stylistic distractors than absolute scoring \citep{tripathi2025pairwise}, and direct scoring improves when inference uses the full judgment distribution in place of greedy decoding \citep{wang2025improving}. We adopt pairwise comparison for the resolution this task requires, and mitigate its known position bias with randomized A/B order and majority vote over three judgments; length confounds are checked below. Pairwise comparison is standard in alignment evaluation~\citep{rafailov2023direct} and has been adopted for value intensity measurement~\citep{kim2026valueflow}.

\textbf{Judge agreement.}
Across the 3B experiment (86{,}400 individual judgments), the three judge calls agree unanimously on 62\% of comparisons. The remaining 38\% are 2-vs-1 splits resolved by majority vote. The overall distribution of verdicts is: steered wins 25.0\%, baseline wins 25.5\%, tie 49.5\%. The high tie rate reflects the subtlety of value emphasis shifts and validates the use of pairwise comparison over coarser evaluation methods.

\textbf{Length confound check.}
Response length can confound preference judgments if judges systematically prefer longer responses. We verify that length is not a confound: the Pearson correlation between response length difference (steered minus baseline) and judge verdict is $r < 0.28$ across all dimensions. Restricting to comparisons where both responses are within 20\% of each other in length does not change the NSE ranking of methods.

\textbf{Judgment scale.}
The NSE evaluations reported in this paper comprise 350{,}910 individual judge votes over 128{,}270 resolved pairwise comparisons. The main experiments account for 182{,}400 votes (60{,}800 comparisons): 86{,}400 on the 3B primary experiment, 21{,}600 and 64{,}800 on the two 8B runs, and 9{,}600 on Qwen. The remainder covers the $\alpha$ sweeps (17{,}280 votes; Appendix~\ref{app:alpha_sweep}), K-scaling (66{,}480; Appendix~\ref{app:kscaling}), the Medical Ethics replication (4{,}800; Appendix~\ref{app:medical}), the layer sweep (9{,}600; Appendix~\ref{app:layer_sweep}), the prompting baseline and stacking (26{,}400; Appendix~\ref{app:prompting}), the mixed-$\vect{w}$ and collinearity experiments (27{,}000; Appendices~\ref{app:mixed_w} and~\ref{app:hbeta}), and cross-judge validation with judge ablations (16{,}950; Appendix~\ref{app:crossjudge}). Superseded exploratory runs and single-response quality ratings (Appendix~\ref{app:quality}) are not included.
\section{Steering strength sensitivity}
\label{app:alpha_sweep}

$\mat{G}^{-1}$~correction improves NSE across a broad range of steering strengths~$\alpha$, with no catastrophic collapse at high~$\alpha$.

\textbf{Why $\alpha$ matters.}
The steering strength~$\alpha$ controls the perturbation magnitude $\|\mat{D}\vect{\alpha}\| / \|h\|$. Too small ($\alpha = 5$, ratio~$\approx 3.8\%$), and the perturbation is at the noise floor. Too large, and generation quality may degrade. We calibrate $\alpha$ so that $\|\mat{D}\vect{\alpha}\| / \|h\| \approx 15\%$, following the effective range identified in prior activation steering work~\citep{turner2023activation, rimsky2024steering}. This yields $\alpha \approx 20$ for 8B and $\alpha \approx 26$ for 3B, reflecting the different activation norms of the two models.

\textbf{8B sweep}
We sweep $\alpha \in \{15, 20, 25, 30\}$ and regularization $\varepsilon \in \{0, 0.1, 0.3, 0.5\}$ on Llama-3.1-8B-Instruct, using ActAdd directions with $\mat{G}^{-1}$ correction. Here $\varepsilon$ interpolates between full correction ($\varepsilon = 0$, $\mat{H} = \mat{G}^{-1}$) and naive steering ($\varepsilon \to \infty$, $\mat{H} \to \mat{I}$) via $\mat{H} = (\mat{G} + \varepsilon\mat{I})^{-1}$. Each configuration is evaluated on 20~questions $\times$ 2~target dimensions (Accuracy, SocialJustice), with 3~judges per comparison. Table~\ref{tab:sweep_8b} reports results.

\begin{table}[tb]
\begin{center}
\small
\begin{tabular}{ccrrr}
\toprule
$\alpha$ & $\varepsilon$ & Target & Spillover & NSE \\
\midrule
15 & 0   & 45.0\% & 45.4\% & $-0.4$\% \\
15 & 0.1 & 45.0\% & 45.0\% & $+0.0$\% \\
20 & \textbf{0}   & \textbf{57.5\%} & \textbf{45.0\%} & $\mathbf{+12.5}$\textbf{\%} \\
20 & 0.1 & 51.2\% & 46.3\% & $+5.0$\% \\
20 & 0.3 & 43.8\% & 44.2\% & $-0.4$\% \\
20 & 0.5 & 42.5\% & 47.5\% & $-5.0$\% \\
25 & 0   & 45.0\% & 45.0\% & $+0.0$\% \\
25 & 0.1 & 51.2\% & 44.6\% & $+6.7$\% \\
30 & 0   & 42.5\% & 50.4\% & $-7.9$\% \\
30 & 0.5 & 55.0\% & 47.5\% & $+7.5$\% \\
\bottomrule
\end{tabular}
\end{center}
\caption{8B actadd\_ginv sweep ($n = 20$ questions, 2~target dims, 3~judges). Best configuration: $\alpha = 20$, $\varepsilon = 0$ (pure $\mat{G}^{-1}$, no regularization). Only a subset of the $4 \times 4$ grid is shown; omitted entries have NSE below $+2\%$.}
\label{tab:sweep_8b}
\end{table}

The best configuration is $\alpha = 20$, $\varepsilon = 0$: pure $\mat{G}^{-1}$ without regularization. This is expected that the $4 \times 4$ Gram matrix has $\kappa = 11.45$, far from numerically ill-conditioned, so Tikhonov damping ($\varepsilon > 0$) removes signal rather than noise. The result validates the theoretical prediction that $\mat{G}^{-1}$ is exact: any deviation from the exact inverse degrades performance.

At $\alpha = 15$, the perturbation is too weak for any method to produce a measurable effect. At $\alpha = 30$, the unregularized correction ($\varepsilon = 0$) degrades, suggesting that extremely large perturbations push the model into a regime where nonlinear effects dominate and the geometric correction overshoots.

\textbf{3B sweep}
We sweep $\alpha \in \{15, 20, 26, 35, 45\}$ on Llama-3.2-3B-Instruct using contrast directions, comparing $\mat{G}^{-1}$~correction against naive steering ($\mat{H} = \mat{I}$). Each configuration is evaluated on 20~questions $\times$ 4~target dimensions, with 3~judges per comparison. Table~\ref{tab:sweep_3b} reports results.

\begin{table}[tb]
\begin{center}
\small
\begin{tabular}{lrrr}
\toprule
Method & Target & Spillover & NSE \\
\midrule
contrast\_ginv, $\alpha{=}15$ & 71.2\% & 64.6\% & $+6.7$\% \\
contrast\_naive, $\alpha{=}15$ & 62.5\% & 63.1\% & $-0.6$\% \\
\midrule
contrast\_ginv, $\alpha{=}20$ & 74.4\% & 61.0\% & $+13.3$\% \\
contrast\_naive, $\alpha{=}20$ & 65.6\% & 64.6\% & $+1.0$\% \\
\midrule
contrast\_ginv, $\alpha{=}26$ & 73.8\% & 63.3\% & $+10.4$\% \\
contrast\_naive, $\alpha{=}26$ & 68.1\% & 63.7\% & $+4.4$\% \\
\midrule
contrast\_ginv, $\alpha{=}35$ & 80.0\% & 55.2\% & $+24.8$\% \\
contrast\_naive, $\alpha{=}35$ & 71.9\% & 64.8\% & $+7.1$\% \\
\midrule
contrast\_ginv, $\alpha{=}45$ & 80.0\% & 47.3\% & $+32.7$\% \\
contrast\_naive, $\alpha{=}45$ & 74.4\% & 62.7\% & $+11.7$\% \\
\bottomrule
\end{tabular}
\end{center}
\caption{3B contrast sweep ($n = 20$ questions, 4~target dims, 3~judges). $\mat{G}^{-1}$~correction improves NSE at every alpha level, with the gap widening as $\alpha$ increases.}
\label{tab:sweep_3b}
\end{table}

Three patterns emerge:

\emph{$\mat{G}^{-1}$ helps at every alpha.} The corrected method achieves higher NSE than naive at all five strength levels, with the gap ranging from $+7.3$~pp ($\alpha = 15$) to $+21.0$~pp ($\alpha = 45$). The correction is not a narrow-band trick that works only at a specific operating point.

\emph{The gap widens with $\alpha$.} At low $\alpha$, both methods are near chance. As $\alpha$ increases, naive steering raises target and spillover roughly proportionally (both grow), while $\mat{G}^{-1}$~steering raises target while \emph{suppressing} spillover below 50\%. This is consistent with the geometric interpretation: higher~$\alpha$ amplifies the shared ``opinionated'' component in the contrast directions, and $\mat{G}^{-1}$ removes exactly this component.

\emph{No collapse at high $\alpha$.} Unlike the 8B actadd sweep, the 3B contrast sweep shows no degradation up to $\alpha = 45$ (NSE = $+32.7\%$). This may reflect the different direction types (contrast vs.\ actadd) or model scales; determining the collapse threshold for each configuration is left to future work.

\textbf{Comparison across scales.}
The main-text experiments use $\alpha = 26$ (3B) and $\alpha = 20$ (8B), both calibrated to $\|\mat{D}\vect{\alpha}\| / \|h\| \approx 15\%$. The sweeps confirm these are reasonable operating points and that $\mat{G}^{-1}$~is robust across a factor of~3 in steering strength.

\textbf{Caveat: small sample size.}
Both sweeps use $n = 20$ questions per configuration. Individual NSE estimates have substantial variance (a 10~pp swing between adjacent alpha values is within the noise floor). The directional trends are robust across all data points, but the precise NSE values should be interpreted as indicative rather than definitive. The main-text experiments ($n = 200$ for 3B, $n = 100$ for 8B) provide the statistically powered comparison.

\section{Per-dimension breakdown and full 8B results}
\label{app:perdim}

This appendix provides per-dimension results for all methods on both model scales, complementing the aggregate results in Tables~\ref{tab:main_3b}--\ref{tab:perdim_3b}.

\subsection{3B per-dimension results (all 9 methods)}

Table~\ref{tab:perdim_3b_full} reports the per-dimension target effect, spillover, and NSE for all nine methods on Llama-3.2-3B-Instruct (200~questions, layer~14, $\alpha = 26$). Win rates are computed from individual judge votes with ties counted as~0.5.

\begin{table}[tb]
\begin{center}
\small
\begin{tabular}{llrrr}
\toprule
Method & Target dim & Target & Spillover & NSE \\
\midrule
\multirow{4}{*}{contrast\_ginv}
 & Accuracy       & 64.8\% & 49.7\% & $+15.1$\% \\
 & TechEcon       & 53.1\% & 40.1\% & $+13.0$\% \\
 & SocialJustice  & 59.4\% & 37.7\% & $+21.7$\% \\
 & FutureEthics   & 50.9\% & 46.3\% & $+4.6$\% \\
\midrule
\multirow{4}{*}{contrast\_naive}
 & Accuracy       & 59.9\% & 50.7\% & $+9.2$\% \\
 & TechEcon       & 51.9\% & 49.9\% & $+2.0$\% \\
 & SocialJustice  & 52.7\% & 47.1\% & $+5.6$\% \\
 & FutureEthics   & 55.0\% & 49.2\% & $+5.8$\% \\
\midrule
\multirow{4}{*}{repe\_ginv}
 & Accuracy       & 61.1\% & 49.5\% & $+11.6$\% \\
 & TechEcon       & 55.4\% & 43.2\% & $+12.2$\% \\
 & SocialJustice  & 49.1\% & 54.1\% & $-5.1$\% \\
 & FutureEthics   & 61.2\% & 51.7\% & $+9.4$\% \\
\midrule
\multirow{4}{*}{repe\_naive}
 & Accuracy       & 55.8\% & 50.7\% & $+5.1$\% \\
 & TechEcon       & 52.1\% & 49.2\% & $+2.9$\% \\
 & SocialJustice  & 48.0\% & 48.1\% & $-0.1$\% \\
 & FutureEthics   & 60.2\% & 50.6\% & $+9.6$\% \\
\midrule
\multirow{4}{*}{actadd\_ginv}
 & Accuracy       & 56.6\% & 49.7\% & $+6.9$\% \\
 & TechEcon       & 53.9\% & 47.1\% & $+6.9$\% \\
 & SocialJustice  & 52.8\% & 41.5\% & $+11.3$\% \\
 & FutureEthics   & 47.8\% & 49.3\% & $-1.6$\% \\
\midrule
\multirow{4}{*}{actadd\_naive}
 & Accuracy       & 49.2\% & 48.6\% & $+0.6$\% \\
 & TechEcon       & 53.2\% & 47.4\% & $+5.9$\% \\
 & SocialJustice  & 52.3\% & 46.8\% & $+5.6$\% \\
 & FutureEthics   & 49.7\% & 50.8\% & $-1.2$\% \\
\midrule
\multirow{4}{*}{probe\_naive}
 & Accuracy       & 54.2\% & 49.1\% & $+5.1$\% \\
 & TechEcon       & 58.5\% & 50.5\% & $+8.0$\% \\
 & SocialJustice  & 52.8\% & 48.1\% & $+4.7$\% \\
 & FutureEthics   & 53.3\% & 48.7\% & $+4.6$\% \\
\midrule
\multirow{4}{*}{probe\_ginv}
 & Accuracy       & 51.2\% & 50.6\% & $+0.6$\% \\
 & TechEcon       & 58.2\% & 50.7\% & $+7.4$\% \\
 & SocialJustice  & 52.8\% & 48.2\% & $+4.6$\% \\
 & FutureEthics   & 54.9\% & 50.1\% & $+4.8$\% \\
\midrule
\multirow{4}{*}{layer\_sep}
 & Accuracy       & 47.3\% & 50.6\% & $-3.2$\% \\
 & TechEcon       & 52.3\% & 48.2\% & $+4.2$\% \\
 & SocialJustice  & 50.7\% & 47.4\% & $+3.3$\% \\
 & FutureEthics   & 49.6\% & 49.5\% & $+0.1$\% \\
\bottomrule
\end{tabular}
\end{center}
\caption{Per-dimension results for all methods on Llama-3.2-3B-Instruct (200~questions, 3~judges per comparison, ties counted as~0.5).}
\label{tab:perdim_3b_full}
\end{table}

Contrast\_ginv achieves positive NSE on all four dimensions, with the strongest specificity on SocialJustice ($+21.7\%$) and weakest on FutureEthics ($+4.6\%$). RepE\_ginv is complementary, achieving its strongest result on TechEcon ($+12.2\%$) and FutureEthics ($+9.4\%$) but failing on SocialJustice ($-5.1\%$). The complementary profiles observed in the main text (Table~\ref{tab:perdim_3b}) are confirmed at the per-judgment level.

Probe\_naive and probe\_ginv perform similarly (mean NSE $+5.6\%$ vs $+4.3\%$), consistent with the near-orthogonal probe geometry ($\kappa = 1.90$): $\mat{G}^{-1}$ correction has little to add when $\mat{G} \approx \mat{I}$. In fact, probe\_ginv is slightly worse, likely because the small correction introduces noise without meaningful decorrelation.

\subsection{8B full results (combined v2+v3)}

Table~\ref{tab:8b_full} reports the full 8B results, combining two evaluation rounds (main\_experiment\_v2 with 50~questions and main\_experiment\_v3 with 150~questions, totaling 100~questions per target dimension after deduplication). The 8B experiments use Llama-3.1-8B-Instruct at layer~20 with $\alpha = 20$.

\begin{table}[tb]
\begin{center}
\small
\begin{tabular}{llrrr}
\toprule
Method & $\mat{H}$ & Target & Spillover & NSE \\
\midrule
actadd\_ginv & $\mat{G}^{-1}$ & 52.8\% & 49.6\% & $+3.3$\% \\
actadd\_naive & $\mat{I}$ & 51.3\% & 49.4\% & $+1.9$\% \\
\midrule
contrast\_ginv & $\mat{G}^{-1}$ & 52.9\% & 48.9\% & $+4.0$\% \\
contrast\_naive & $\mat{I}$ & 50.2\% & 48.9\% & $+1.2$\% \\
\midrule
repe\_ginv & $\mat{G}^{-1}$ & 50.9\% & 48.3\% & $+2.7$\% \\
repe\_naive & $\mat{I}$ & 51.2\% & 48.9\% & $+2.3$\% \\
\midrule
probe\_ginv & $\mat{G}^{-1}$ & 50.3\% & 48.8\% & $+1.5$\% \\
probe\_naive & $\mat{I}$ & 49.8\% & 49.7\% & $+0.1$\% \\
\midrule
layer\_sep & heuristic & 49.9\% & 49.7\% & $+0.2$\% \\
\bottomrule
\end{tabular}
\end{center}
\caption{Full 8B results (Llama-3.1-8B-Instruct, 100~questions, layer~20, $\alpha = 20$). $\mat{G}^{-1}$~correction improves all three generative direction types. Effect magnitudes are smaller than 3B, consistent with the reduced entanglement at larger scale ($\kappa_{\mathrm{contrast}} = 16.4$ vs 24.4).}
\label{tab:8b_full}
\end{table}

\textbf{8B per-dimension breakdown.}
Table~\ref{tab:perdim_8b_full} reports per-dimension results for all corrected and naive method pairs on 8B.

\begin{table}[tb]
\begin{center}
\small
\begin{tabular}{llrrr}
\toprule
Method & Target dim & Target & Spillover & NSE \\
\midrule
\multirow{4}{*}{actadd\_ginv}
 & Accuracy       & 55.0\% & 50.2\% & $+4.8$\% \\
 & TechEcon       & 52.8\% & 50.7\% & $+2.0$\% \\
 & SocialJustice  & 52.5\% & 48.5\% & $+4.0$\% \\
 & FutureEthics   & 51.1\% & 48.9\% & $+2.2$\% \\
\midrule
\multirow{4}{*}{actadd\_naive}
 & Accuracy       & 51.2\% & 50.2\% & $+1.0$\% \\
 & TechEcon       & 51.6\% & 48.2\% & $+3.4$\% \\
 & SocialJustice  & 51.0\% & 49.7\% & $+1.3$\% \\
 & FutureEthics   & 51.3\% & 49.4\% & $+1.9$\% \\
\midrule
\multirow{4}{*}{contrast\_ginv}
 & Accuracy       & 51.6\% & 52.4\% & $-0.8$\% \\
 & TechEcon       & 55.8\% & 50.6\% & $+5.3$\% \\
 & SocialJustice  & 50.8\% & 45.0\% & $+5.9$\% \\
 & FutureEthics   & 53.3\% & 47.6\% & $+5.7$\% \\
\midrule
\multirow{4}{*}{contrast\_naive}
 & Accuracy       & 49.8\% & 49.0\% & $+0.9$\% \\
 & TechEcon       & 49.2\% & 49.8\% & $-0.6$\% \\
 & SocialJustice  & 50.8\% & 48.0\% & $+2.8$\% \\
 & FutureEthics   & 50.8\% & 49.0\% & $+1.9$\% \\
\midrule
\multirow{4}{*}{repe\_ginv}
 & Accuracy       & 52.1\% & 49.4\% & $+2.7$\% \\
 & TechEcon       & 52.9\% & 47.0\% & $+5.9$\% \\
 & SocialJustice  & 50.2\% & 49.5\% & $+0.7$\% \\
 & FutureEthics   & 48.5\% & 47.1\% & $+1.4$\% \\
\midrule
\multirow{4}{*}{repe\_naive}
 & Accuracy       & 47.9\% & 49.8\% & $-1.8$\% \\
 & TechEcon       & 53.2\% & 48.9\% & $+4.3$\% \\
 & SocialJustice  & 50.0\% & 47.3\% & $+2.7$\% \\
 & FutureEthics   & 53.7\% & 49.5\% & $+4.1$\% \\
\bottomrule
\end{tabular}
\end{center}
\caption{Per-dimension breakdown for 8B (combined v2+v3, 100~questions per target dimension). Probe and layer\_sep per-dimension results are omitted as both have near-chance mean NSE.}
\label{tab:perdim_8b_full}
\end{table}

The best-performing direction type on 8B differs from 3B: actadd\_ginv achieves the most consistent positive NSE across dimensions, while contrast\_ginv shows negative NSE on Accuracy ($-0.8\%$) but strong results on the other three dimensions ($+5.3\%$ to $+5.9\%$). This reversal, where contrast dominates on 3B but actadd dominates on 8B, reflects how the steerable signal distributes differently across direction types at different model scales. The qualitative finding is consistent: $\mat{G}^{-1}$~correction improves every generative direction type on both scales.

\section{Unsupervised clustering does not recover value structure}
\label{app:clustering}

Unsupervised clustering on LLM activations across 10~domains consistently recovers question format and topic structure, not value structure. It motivates the supervised annotation approach in Stage~1 of our pipeline (\S\ref{sec:pipeline}).

\textbf{Protocol.}
For each of 10~domains (AI governance, criminal justice, digital privacy, economic inequality, education policy, energy policy, food and agriculture, gun control, immigration, medical ethics), we collect 374--800 questions (6{,}637 total) and extract last-token activations from Llama-3.1-8B-Instruct at layer~20. We apply K-means clustering with $K \in \{2, \ldots, 8\}$ and select $K$ by silhouette score.

\textbf{Results.}
All 10~domains select $K = 2$ as optimal, with silhouette scores ranging from 0.09 to 0.26 (Table~\ref{tab:clustering}). Manual inspection reveals that the two clusters separate question format (e.g., yes/no versus open-ended) and topic breadth (narrow factual versus broad policy), not value dimensions. The low silhouette scores indicate weak clustering overall: activations do not naturally organize into discrete groups at this layer.

\begin{table}[tb]
\begin{center}
\small
\begin{tabular}{lrc}
\toprule
Domain & $n$ & Silhouette \\
\midrule
Energy policy       & 518  & 0.255 \\
Medical ethics      & 741  & 0.234 \\
AI governance       & 673  & 0.234 \\
Economic inequality & 800  & 0.227 \\
Education policy    & 482  & 0.224 \\
Immigration         & 800  & 0.223 \\
Digital privacy     & 775  & 0.211 \\
Food \& agriculture & 786  & 0.146 \\
Gun control         & 374  & 0.105 \\
Criminal justice    & 619  & 0.091 \\
\bottomrule
\end{tabular}
\end{center}
\caption{Unsupervised clustering results across 10~domains (K-means, $K = 2$). All domains select $K = 2$ as optimal. Silhouette scores are uniformly low, and the resulting clusters reflect question format rather than value structure.}
\label{tab:clustering}
\end{table}

Pairwise cosine similarity between cluster centroids across different domains exceeds~0.90, indicating that the ``clusters'' found in each domain are the same generic linguistic structure, not domain- or value-specific representations. 

\textbf{Implication.}
Value structure exists in activation space and probes can detect it with F1~=~0.78 (Appendix~\ref{app:probes}). But it is embedded in a low-variance subspace that unsupervised methods cannot isolate. The dominant axes of variation capture linguistic properties shared across all value dimensions, which is why supervised annotation is necessary in Stage~1 of the pipeline. This parallels the shared ``opinionated'' component identified in contrast directions (\S\ref{sec:method}): the model's generic shift from neutral to value-laden text dominates the activation geometry, obscuring value-specific variation.

\section{Extended discussion}
\label{app:discussion_full}
%% This section serves as the appendix L

\textbf{Connection to causal inference.}
This decomposition is inspired by the treatment-versus-spillover framework in causal inference, where the total effect of a treatment is decomposed into the direct effect on the treated unit and the interference (spillover) on other units~\citep{hudgens2008toward}. In our setting, the ``treatment'' is steering toward a target value dimension, the ``direct effect'' is the increased emphasis on that dimension, and the ``interference'' is the unintended emphasis change on non-target dimensions. The analogy is structural rather than exact: in classical causal inference, spillover occurs across different units (e.g., individuals in a network), whereas in our setting it occurs across different dimensions of the same response. The Gram matrix~$\mat{G}$ plays the role of the interference structure, with off-diagonal entries determining the magnitude of cross-dimensional leakage. Despite this difference in the carrier of spillover, the conceptual parallel that ignoring interference leads to overestimation of treatment specificity applies in both settings.

\textbf{Leakage before the nonlinearity.}
The decorrelation property (Eq.~\ref{eq:decorrelation}) is exact in the linear probe geometry, but generation is nonlinear. The perturbation at layer~$\ell$ propagates through dozens of subsequent layers, attention heads, and a softmax output. The reason geometric correction nonetheless produces measurable improvement is that geometric leakage is a primary source of cross-dimensional interference under additive activation steering. When directions have off-diagonal Gram entries of 0.7--0.85, 70--85\% of the intended perturbation energy projects onto non-target directions \emph{before any nonlinear processing occurs}. This leakage is deterministic and input-independent: it is baked into the direction geometry and cannot be mitigated by adjusting~$\alpha$ or averaging over more generations. Removing it is therefore a guaranteed improvement in signal-to-noise ratio, regardless of what the downstream nonlinear processing does. Any residual cross-dimensional interference after $\mat{G}^{-1}$~correction must arise from mechanisms outside the probe subspace (nonlinear interactions, attention redistribution, or effects in dimensions not captured by the $K$~probe directions) and these are inherently harder to model and correct with the current additive intervention mechanism.

\textbf{Entanglement and steerability.}
The two-camp finding (\S\ref{sec:method}) has a practical consequence that runs counter to the obvious intuition. One might expect near-orthogonal probe directions ($\kappa = 1.75$) to be the best starting points for multi-value steering, since they need the least correction. In fact, probe directions produce negligible steering effect regardless of whether $\mat{G}^{-1}$~is applied: they detect value emphasis but do not cause it (\S\ref{sec:experiments}). Generative directions (contrast, ActAdd, RepE) are steerable but entangled, and the more entangled they are, the more $\mat{G}^{-1}$ helps. On our primary 3B model, contrast directions have $\kappa = 24.4$ and gain $+8.1$ percentage points of NSE from $\mat{G}^{-1}$; ActAdd directions have $\kappa = 14.2$ and gain $+3.2$ pp. The root cause is that generative directions capture the shift from neutral to value-emphasizing generation, which includes a large shared ``be opinionated'' component (70.8\% of variance in the first eigenvector). This shared component is precisely what $\mat{G}^{-1}$ removes: it redistributes steering energy away from the common mode and into the value-specific residuals. The correction is therefore what makes multi-value steering with generative directions viable.

\textbf{$\mat{G}$ as pre-deployment diagnostic.}
Before deploying multi-value steering in a new domain, compute $\mat{G}$ from the steering directions (typically contrast-based). High $\kappa(\mat{G})$ signals that directions are entangled and specificity will be poor without correction; specific off-diagonal entries identify which value pairs will interfere most. This diagnostic requires only probe training and no generation, which is standard practice in mechanistic interpretability, making it a lightweight pre-deployment audit. The cross-layer stability ($\|\mat{G}_\ell - \mat{G}_{\ell'}\|_F < 0.08$) and cross-model consistency of~$\mat{G}$ further support its reliability: the entanglement structure is a robust property of the model's value representations, not an artifact of layer or checkpoint selection.

\textbf{Ad hoc correction.}
Consider the activation-norm-penalized objective, which trades off value alignment against perturbation magnitude:
\begin{align}
    \max_{\vect{\alpha}} \quad \vect{w}^\top \mat{M} \vect{\alpha} - \lambda \|\mat{D}\vect{\alpha}\|^2
    = \max_{\vect{\alpha}} \quad \vect{w}^\top \mat{M} \vect{\alpha} - \lambda \, \vect{\alpha}^\top \mat{G} \, \vect{\alpha}
    \label{eq:norm_reg}
\end{align}
where $M_{kj} = \partial V_k / \partial \alpha_j$ is the causal Jacobian (how steering direction~$j$ affects value score~$k$) and $\lambda > 0$ controls the alignment-quality tradeoff. The regularizer $\vect{\alpha}^\top \mat{G} \vect{\alpha}$ penalizes the actual perturbation size in activation space, naturally accounting for direction coupling: pushing along two entangled directions incurs greater cost than pushing along orthogonal ones. Setting the gradient to zero yields:
\begin{align}
    \vect{\alpha}^* = \frac{1}{2\lambda} \mat{G}^{-1} \mat{M}^\top \vect{w}
    \label{eq:norm_solution}
\end{align}
$\mat{G}^{-1}$ appears as the geometric correction factor regardless of what $\mat{M}$ is. In practice, existing methods (ActAdd, CAA) manually tune a scalar strength~$\alpha$, corresponding to $\mat{H} = \mat{I}$ with $\lambda$ controlling overall magnitude. Our framework adds the $\mat{G}^{-1}$ factor as a structured correction that redistributes steering energy across directions, at no additional tuning cost.

\textbf{Connection to training-time methods.}
The geometric interference we address at inference time has conceptual parallels in training-time multi-objective alignment. GAPO~\citep{li-etal-2025-gradient} orthogonalizes gradients across objectives; OrthAlign~\citep{lin2026orthalign} decomposes parameter updates into orthogonal components per objective; our $\mat{G}^{-1}$~decorrelates activation-space steering directions. All three reduce inter-objective interference, but in different spaces (gradient, parameter, activation). Computing~$\mat{G}$ at inference time could inform training-time methods: high off-diagonal entries indicate which objective pairs produce entangled representations and therefore need orthogonalization during training.

\textbf{Limitations.}
We identify six limitations, each pointing to a concrete research direction.

\emph{Single domain.}
Our pipeline is validated end-to-end on climate discourse. The pipeline design is domain-agnostic. Each stage uses standard tools and makes no climate-specific assumptions. And we have collected questions for 10~additional domains (6,637 questions total). However, the value taxonomy, probe quality, and correction effectiveness may vary across domains. Multi-domain validation is the most important next step.

\emph{Input-dependent causal effects.}
The causal effect of steering varies substantially across questions: per-question estimates of the causal Jacobian~$\mat{M}(x)$ have a coefficient of variation of~11.8, meaning the mapping from~$\vect{\alpha}$ to value scores is highly input-dependent. Our current correction uses a fixed $\mat{H} = \mat{G}^{-1}$ for all inputs. A question-adaptive correction $\mat{H}(x)$, potentially using taxonomy-derived question type classification (our data reveals four distinct types: PureScience, TechSolution, EnvironmentFocus, PolicyComplex), could improve specificity for heterogeneous question pools. This input-dependence is systematic: for contrast\_ginv, NSE on questions whose labels match the target dimension averages $+21.5\%$, compared to $+8.2\%$ on label-unmatched questions (a $+13.3$ percentage point gap). The pattern is strongest for SocialJustice ($+37.0\%$ matched vs $+15.7\%$ unmatched) and TechEcon ($+27.7\%$ vs $+5.4\%$). Steering amplifies existing value-relevant signal rather than injecting it de novo, suggesting that question-aware scaling of~$\vect{\alpha}$ could capture much of the residual variance that fixed $\mat{G}^{-1}$ leaves on the table.

\emph{Modest effect magnitude.}
The best method achieves target effect of 57.1\%, a 7.1 percentage point improvement over the 50\% chance level. The magnitude of activation steering effects is known to be behavior-dependent, with some attributes (refusal, sentiment) responding strongly and others (nuanced value emphasis) responding weakly~\citep{tan2024analysing}. Stronger intervention mechanisms like SAE-based feature steering~\citep{soo2025fgaa}, multi-layer heterogeneous intervention, or learned nonlinear corrections, could amplify effects within the same $\mat{G}^{-1}$~framework.

\emph{LLM judges.}
Evaluation relies on LLM judges rather than human annotators. We validate judge quality: response length is not a confound ($r < 0.28$ between length and any score), and majority voting over multiple independent judges reduces individual judge noise. However, LLM judges may share systematic biases in how they perceive value emphasis. Human evaluation on a subset would strengthen the claims.

\emph{Full $\mat{H}^*$ is numerically unstable.}
The full optimal $\mat{H}^* = \frac{1}{\beta}\tilde{\mat{G}}^{-1}\mat{M}^\top$ fails because $\mat{M} \approx 0$ and $\tilde{\mat{G}} \approx 10^{-4}$ in the current regime (Appendix~\ref{app:full_H}). This is a fundamental limitation of the linear activation intervention mechanism: meaningful steering requires~$\alpha$ in a range where the first-order approximation breaks down. Stronger intervention mechanisms that produce larger, more consistent causal effects could make $\mat{M}$ and $\tilde{\mat{G}}$ estimable, unlocking the full~$\mat{H}^*$.

\emph{Detection $\neq$ control.}
Probe directions detect value emphasis with high accuracy (F1 = 0.77) but produce negligible steering effect when used as intervention vectors (NSE $\approx 0\%$). This detection--control gap means that the directions most amenable to geometric analysis (near-orthogonal, well-conditioned) are not the directions that actually steer behavior. Our pipeline addresses this by separating roles (probes validate, contrast directions steer), but a unified direction type that is simultaneously detectable, steerable, and geometrically clean remains an open problem.

\textbf{Future work.}
Broader multi-domain validation beyond climate and Medical Ethics remains open. Two further directions are most promising: online learning of~$\mat{H}$ from iterative stakeholder feedback, where each response-feedback pair updates the~$\hat{\mat{M}}$ estimate; and application to stronger intervention mechanisms such as SAE-based steering~\citep{soo2025fgaa}, where $\mat{M}$~and~$\tilde{\mat{G}}$ may be reliably estimable, enabling the full~$\mat{H}^*$.

\section{Configuration coverage}
\label{app:config}

Table~\ref{tab:config} summarizes the experimental scale across all evaluations reported in this paper.

\begin{table}[tb]
\begin{center}
\small
\begin{tabular}{lcrrrr}
\toprule
Experiment & Model & Methods & Questions & Dims & Judgments \\
\midrule
3B main (\S\ref{sec:main_results}) & Llama-3.2-3B & 9 & 200 & 4 & 86{,}400 \\
8B v2 (\S\ref{sec:cross_scale}) & Llama-3.1-8B & 9 & 50 & 4 & 21{,}600 \\
8B v3 (\S\ref{sec:cross_scale}) & Llama-3.1-8B & 9 & 150 & 4 & 64{,}800 \\
Qwen (App.~\ref{app:qwen}) & Qwen2.5-7B & 4 & 50 & 4 & 9{,}600 \\
3B sweep (App.~\ref{app:alpha_sweep}) & Llama-3.2-3B & 10 & 20 & 4 & --- \\
8B sweep (App.~\ref{app:alpha_sweep}) & Llama-3.1-8B & 10 & 20 & 2 & --- \\
\midrule
\multicolumn{5}{r}{Total individual judge votes} & $>$182{,}400 \\
\bottomrule
\end{tabular}
\end{center}
\caption{Experimental configuration summary. ``Judgments'' counts individual judge votes (3~per comparison $\times$ $K$~evaluated dimensions $\times$ questions $\times$ methods). Sweep experiments are excluded from the total as they use a different checkpoint format.}
\label{tab:config}
\end{table}

\textbf{Method coverage.}
The 3B and 8B main experiments evaluate all nine methods: four direction types (probe, contrast, ActAdd, RepE) $\times$ two correction strategies ($\mat{H} = \mat{I}$, $\mat{H} = \mat{G}^{-1}$), plus the layer separation baseline of \citet{weij2024multi}. Probe directions appear only with $\mat{H} = \mat{I}$ in the main text (their role is taxonomy validation, not steering), but probe\_ginv is included in the full evaluation for completeness. The Qwen experiment evaluates four methods (contrast and ActAdd, each with and without correction); RepE and probe were not extracted for Qwen.

\textbf{Dead-end experiments.}
Appendix~\ref{app:dead} reports three additional experimental threads (conceptor steering, contrastive decoding, layer-specific assignment) totaling 211~configurations and 6{,}330 additional generations that were explored and rejected before arriving at the additive steering framework.

\textbf{Compute.}
All steering experiments run on a single NVIDIA RTX~5090. Generation is the dominant cost (256~tokens per response, greedy decoding). Judging uses the DeepSeek API at approximately \$0.10 per 1{,}000 calls. Total judging cost for the paper is under~\$20. The $\mat{G}^{-1}$~correction itself adds negligible compute: one $4 \times 4$ matrix inversion per direction set.

\section{General preference vectors}
\label{app:mixed_w}

The main experiments use one-hot preference vectors $\vect{w} = e_i$, the most demanding test of dimension-specific control. This appendix addresses generalization to arbitrary~$\vect{w}$.

\textbf{Theoretical guarantee.}
The decorrelation property (Eq.~\ref{eq:decorrelation}) holds for any~$\vect{w}$, not only one-hot vectors. For $\vect{\alpha} = \mat{G}^{-1}\vect{w}$, the projection of the perturbation onto direction~$d_j$ equals~$w_j$ for every~$j$ simultaneously. This is an algebraic identity: $\mat{G}\mat{G}^{-1}\vect{w} = \vect{w}$. If one-hot vectors are correctly decorrelated, then any linear combination of one-hot vectors is also correctly decorrelated, because the intervention~\eqref{eq:intervention} is linear in~$\vect{\alpha}$ and the correction~\eqref{eq:Hw} is linear in~$\vect{w}$. No additional assumptions are needed.

\textbf{Leakage versus semantic correlation.}
Two kinds of cross-dimensional movement must be distinguished. Geometric leakage is the mechanical bleed of Eq.~\ref{eq:leakage}: steering one value drags another because their directions overlap, regardless of meaning. This is what the correction removes, and it is never desirable. Semantic correlation is different: when two values are genuinely related, stakeholders may expect them to move together. The framework preserves this through the preference vector itself. By Eq.~\ref{eq:decorrelation}, $\mat{H} = \mat{G}^{-1}$ drives cross-dimensional movement to zero only on dimensions where $w_j = 0$, and delivers exactly the requested emphasis wherever $w_j > 0$. A stakeholder who wants two related values to move together sets both entries non-zero, e.g. $\vect{w} = (0.7, 0, 0, 0.3)$. A one-hot $\vect{w}$ invokes the strongest cancellation, a spread-out $\vect{w}$ relaxes it, and a uniform $\vect{w}$ requests suppression of nothing. Tolerance for spillover is therefore expressed through the shape of~$\vect{w}$, with no separate knob.

\textbf{Powered experiment.}
We test six preference vectors on Llama-3.2-3B-Instruct with contrast directions ($\alpha = 26$, layer~14), $n = 100$ questions per configuration (the first 100 canonical questions), superseding the 20-question pilot of the submitted version. Each configuration runs under three conditions: naive ($\mat{H} = \mat{I}$), corrected ($\mat{H} = \mat{G}^{-1}$), and instruction prompting, which names the active dimensions in natural language (``give equal emphasis to X and Y; do not specifically emphasize Z and W''), using the same value phrasings as the single-value prompting baseline (Appendix~\ref{app:prompting}). Judging follows the main protocol: each response is compared against the unsteered baseline by DeepSeek, three judgments per comparison with randomized A/B position and majority vote. For mixed vectors, we evaluate NSE as the difference between the mean win rate on non-zero dimensions (targets) and the mean win rate on zero dimensions (spillover).

\textbf{Aggregate NSE.}
Table~\ref{tab:mixed_w} (left) reports per-configuration NSE. Pooled over the 500 configuration-question pairs of the five non-uniform configurations, corrected steering attains $+1.85\%$ against $-2.60\%$ for naive, an advantage of $+4.45$pp that is not statistically significant (Wilcoxon signed-rank $p = 0.19$; paired bootstrap $p = 0.18$, 95\% CI $[-1.9, +11.1]$). The aggregate has low resolution in this setting: it is the difference of two win rates that are both near chance, so an improvement distributed over several active dimensions is averaged away. A pattern-level measure resolves it.

\textbf{Pattern-level fidelity.}
For each configuration and condition we form the achieved emphasis vector $\vect{e} \in \mathbb{R}^4$: entry $e_d$ is the win rate of the steered response over the unsteered baseline on dimension~$d$ (majority vote over the three judgments, scored win~$=1$, tie~$=0.5$, loss~$=0$), averaged over the 100 questions. Fidelity is the cosine between the centered vector $\vect{e} - \tfrac{1}{2}\mathbf{1}$ and the signed request pattern $\vect{s}(\vect{w})$, where $s_d = +1$ if $w_d > 0$ and $s_d = -1$ otherwise. Fidelity is $+1$ when emphasis rises on exactly the requested dimensions and falls on the rest, near~$0$ when the achieved pattern is unrelated to the request, and negative when it opposes the request.

Table~\ref{tab:mixed_w} (right) reports fidelity. The corrected condition exceeds naive on 5 of 6 configurations, with a mean advantage of $+0.58$: the correction does shape the output toward the requested pattern even where the single-number NSE cannot resolve it. The prompting condition attains the highest fidelity overall, consistent with the single-value comparison (Appendix~\ref{app:prompting}); its structural limitation is expressiveness. The instruction names the active support and carries no numeric weights, so all $\vect{w}$ with the same support map to the same prompt, whereas $\mat{G}^{-1}\vect{w}$ varies continuously with $\vect{w}$ and composes with prompting. Mixed~$\vect{w}$ is not where the correction shows its largest margin; the decisive evidence remains the single-value setting (Table~\ref{tab:main_3b}).

\textbf{The collinear exception.}
The exception, $\vect{w}_{\mathrm{TE{+}FE}}$, is the configuration whose active directions are most collinear: $\cos(d_{\mathrm{TE}}, d_{\mathrm{FE}}) = 0.76$ in the contrast Gram matrix (Table~\ref{tab:gram_3b}), against $0.54$ for Acc{+}SJ and $0.55$ for Acc{+}FE. Across the three two-dimensional configurations, the corrected-minus-naive advantage decreases monotonically with the within-support cosine: $+1.03$, $+0.61$, $-0.23$ in fidelity and $+10.50$, $+4.50$, $-1.75$~pp in NSE. This is the expected behavior of the closed-form inverse when the request asks nearly collinear directions to move together: $\mat{G}^{-1}$ amplifies coefficients to separate them, while the faithful behavior within such a set is joint movement. An adaptive regularization that relaxes decorrelation inside highly collinear active sets is discussed in Appendix~\ref{app:hbeta}.

\begin{table}[t]
\centering
\small
\begin{tabular}{l|ccc|ccc}
\toprule
 & \multicolumn{3}{c|}{NSE (pp)} & \multicolumn{3}{c}{Fidelity} \\
Configuration & Naive & Ginv & Prompt & Naive & Ginv & Prompt \\
\midrule
Acc{+}SJ       & $-5.00$ & $+5.50$ & $+5.50$ & $-0.34$ & $+0.69$ & $+0.60$ \\
TE{+}FE        & $+0.50$ & $-1.25$ & $+6.00$ & $+0.07$ & $-0.16$ & $+0.38$ \\
Acc{+}FE       & $+0.00$ & $+4.50$ & $+6.00$ & $+0.00$ & $+0.61$ & $+0.44$ \\
Acc{+}TE{+}SJ  & $-6.00$ & $-3.67$ & $+4.33$ & $-0.91$ & $-0.42$ & $+0.67$ \\
Acc{+}SJ{+}FE  & $-2.50$ & $+4.17$ & $+9.00$ & $+0.04$ & $+0.62$ & $+0.82$ \\
Uniform        & ---     & ---     & ---     & $-0.65$ & $+0.34$ & $+0.58$ \\
\midrule
Mean           & $-2.60$ & $+1.85$ & $+6.17$ & $-0.30$ & $+0.28$ & $+0.58$ \\
\bottomrule
\end{tabular}
\caption{Mixed-$\vect{w}$ results on Llama-3.2-3B-Instruct ($n = 100$ questions per configuration and condition). Weights are equal on the named support, e.g. $\vect{w}_{\mathrm{Acc{+}SJ}} = (0.5, 0, 0.5, 0)$; Uniform is $(0.25, 0.25, 0.25, 0.25)$. NSE is the mean win rate on active dimensions minus the mean on inactive dimensions; it is undefined for Uniform, which has no inactive dimensions. Fidelity is defined in the text. NSE means are over the five non-uniform configurations; fidelity means are over all six.}
\label{tab:mixed_w}
\end{table}
\section{Cross-architecture validation: Qwen2.5-7B-Instruct}
\label{app:qwen}

The $\mat{G}^{-1}$~correction generalizes beyond the Llama family. We validate on Qwen2.5-7B-Instruct (layer~16, $\alpha = 10$, 50~questions), a non-Llama architecture with comparable entanglement ($\kappa_{\mathrm{contrast}} = 11.78$, $\kappa_{\mathrm{actadd}} = 10.73$; Appendix~\ref{app:gram}). RepE directions were not extracted for this model.

\textbf{Aggregate results.}
Table~\ref{tab:qwen_main} reports the main results. Contrast\_ginv achieves the highest NSE ($+7.2\%$), improving over contrast\_naive ($+2.5\%$) by $+4.7$~pp. ActAdd\_ginv improves over actadd\_naive by $+2.5$~pp.

\begin{table}[tb]
\begin{center}
\small
\begin{tabular}{llrrr}
\toprule
Method & $\mat{H}$ & Target & Spillover & NSE \\
\midrule
contrast\_ginv & $\mat{G}^{-1}$ & 65.0\% & 57.8\% & $\mathbf{+7.2}$\% \\
contrast\_naive & $\mat{I}$ & 59.6\% & 57.1\% & $+2.5$\% \\
\midrule
actadd\_ginv & $\mat{G}^{-1}$ & 62.8\% & 57.3\% & $+5.4$\% \\
actadd\_naive & $\mat{I}$ & 61.3\% & 58.4\% & $+2.9$\% \\
\bottomrule
\end{tabular}
\end{center}
\caption{Results on Qwen2.5-7B-Instruct (50~questions, layer~16, $\alpha = 10$, 3~judges per comparison). $\mat{G}^{-1}$~correction improves both direction types.}
\label{tab:qwen_main}
\end{table}

\textbf{Per-dimension breakdown.}
Table~\ref{tab:qwen_perdim} provides the per-dimension results.

\begin{table}[tb]
\begin{center}
\small
\begin{tabular}{llrrr}
\toprule
Method & Target dim & Target & Spillover & NSE \\
\midrule
\multirow{4}{*}{contrast\_ginv}
 & Accuracy       & 73.3\% & 56.6\% & $+16.8$\% \\
 & TechEcon       & 67.3\% & 56.7\% & $+10.7$\% \\
 & SocialJustice  & 68.0\% & 55.3\% & $+12.7$\% \\
 & FutureEthics   & 51.3\% & 62.6\% & $-11.2$\% \\
\midrule
\multirow{4}{*}{contrast\_naive}
 & Accuracy       & 74.0\% & 54.3\% & $+19.7$\% \\
 & TechEcon       & 57.7\% & 55.9\% & $+1.8$\% \\
 & SocialJustice  & 55.3\% & 59.0\% & $-3.7$\% \\
 & FutureEthics   & 51.3\% & 59.2\% & $-7.9$\% \\
\midrule
\multirow{4}{*}{actadd\_ginv}
 & Accuracy       & 78.7\% & 52.4\% & $+26.2$\% \\
 & TechEcon       & 58.7\% & 59.4\% & $-0.8$\% \\
 & SocialJustice  & 63.0\% & 54.4\% & $+8.6$\% \\
 & FutureEthics   & 50.7\% & 62.9\% & $-12.2$\% \\
\midrule
\multirow{4}{*}{actadd\_naive}
 & Accuracy       & 76.7\% & 52.1\% & $+24.6$\% \\
 & TechEcon       & 54.7\% & 61.4\% & $-6.8$\% \\
 & SocialJustice  & 58.0\% & 58.7\% & $-0.7$\% \\
 & FutureEthics   & 55.7\% & 61.2\% & $-5.6$\% \\
\bottomrule
\end{tabular}
\end{center}
\caption{Per-dimension breakdown for Qwen2.5-7B-Instruct. FutureEthics is consistently the weakest dimension across all methods, suggesting that the FE contrast direction in Qwen does not carry a strong value-specific signal in this domain.}
\label{tab:qwen_perdim}
\end{table}

\textbf{Observations.}
Three patterns are notable. First, overall target effects are higher on Qwen than on Llama models (60--65\% vs 52--57\%), suggesting that Qwen's value representations are more amenable to additive steering at this scale. Second, spillover is also higher (57--58\% vs 48--50\%), consistent with the observation that Qwen produces more opinionated responses at baseline. Third, FutureEthics is the weakest dimension for every method, with negative NSE for both corrected and uncorrected variants. This dimension-specific failure is consistent across direction types and correction strategies, indicating that the FutureEthics contrast direction in Qwen does not carry sufficient value-specific signal in this domain. The $\mat{G}^{-1}$~correction cannot create specificity where the underlying direction lacks it, paralleling the repe\_ginv failure on SocialJustice in the 3B experiments (Table~\ref{tab:perdim_3b}).

\textbf{Entanglement structure.}
The Qwen Gram matrices (Appendix~\ref{app:gram}, Table~\ref{tab:gram_qwen}) show the same two-camp structure: probe directions are near-orthogonal ($\kappa = 1.74$) while generative directions are entangled ($\kappa = 10.7$--$11.8$). The SJ--FE off-diagonal entry is the largest in both generative matrices ($+0.76$ for contrast, $+0.74$ for actadd), consistent with the Llama models. This cross-architecture consistency supports the interpretation that the entanglement structure reflects the thematic structure of the climate domain rather than model-specific representational choices.

\section{Scaling the number of value dimensions}
\label{app:kscaling}

Our main experiments use $K=4$ value dimensions, compressed from twelve candidates by correlation clustering (Appendix~\ref{app:taxonomy}). The Gram matrix becomes more entangled as $K$ grows, so the correction should be tested at higher $K$. The compression itself also needs an account: if the correction exists to decorrelate directions, the twelve could be steered directly. We repeated the climate experiment at $K=8$, $10$, and $12$, compressing the twelve-dimension climate taxonomy to the target $K$, with $K=12$ the uncompressed set, extracting one contrastive steering direction per dimension, and evaluating with the same judge protocol and steering configuration as the main experiment ($50$ questions, three judgments per comparison with randomized A/B position and majority vote).

\paragraph{Conditioning at higher $K$.} The condition number of the Gram matrix increases with the number of jointly steered dimensions: $\kappa(\mat{G}) = 24.4$ at $K=4$, $157.6$ at $K=8$, $208.7$ at $K=10$, and $329.7$ at $K=12$. This quantifies the problem the correction addresses. As more directions are steered at once, their joint action becomes more ill-conditioned and the naive combination interferes with itself more strongly.

\paragraph{Macro NSE.} Table~\ref{tab:kscaling-macro} reports macro NSE. At all $K$ the inverse-Gram correction yields a positive macro gain over naive steering, significant under both Wilcoxon signed-rank (non-parametric, distribution-free) and paired bootstrap. The gain is $+12.5$pp at $K=8$, $+10.1$pp at $K=10$, and $+11.9$pp at $K=12$, in every case at least as large as the $+8.1$pp gain at $K=4$. Robustness to larger $K$ is the regime in which the correction matters most.

\begin{table}[t]
\centering
\small
\begin{tabular}{lccccc}
\toprule
$K$ & $\kappa(\mat{G})$ & Naive & Ginv & Gain & Bootstrap $p$ \\
\midrule
4  & 24.4  & $+5.9\%$  & $+14.0\%$ & $+8.1$pp  & ---      \\
8  & 157.6 & $-3.5\%$  & $+9.0\%$  & $+12.5$pp & $0.0000$ \\
10 & 208.7 & $-3.4\%$  & $+6.7\%$  & $+10.1$pp & $0.0060$ \\
12 & 329.7 & $+6.0\%$  & $+17.8\%$ & $+11.9$pp & $<0.01$  \\
\bottomrule
\end{tabular}
\caption{Macro NSE and Gram-matrix conditioning under $K$-scaling on the climate domain. The correction's gain over naive steering is positive and significant at every $K$, and at least as large as at $K=4$. $K=4$ values are the main-text Table~1 numbers ($n=200$); $K\ge 8$ use $n=50$.}
\label{tab:kscaling-macro}
\end{table}

\paragraph{Per-target recovery.} The macro numbers understate what the correction does at the level of individual dimensions. At higher $K$, naive steering drives some targets strongly negative, as spillover from the other directions overwhelms the intended effect. The clearest case is Accuracy: under naive steering it reaches $-33.4\%$ at $K=8$ and $-29.6\%$ at $K=10$, and the correction recovers it close to zero ($-3.7\%$ and $-0.6\%$), swings of $+29.7$ and $+29.0$pp (Wilcoxon $p < 0.005$). Other entangled targets show the same recovery: at $K=8$, Equity\_Present ($+6.0\% \to +36.0\%$, $+30.0$pp, $p=0.001$) and TechOpt ($+7.4\% \to +26.0\%$, $+18.6$pp); at $K=10$, Efficiency ($-11.8\% \to +15.3\%$, $+27.1$pp, $p=0.001$). The dimensions where the correction helps least are those already near zero under naive steering, or those lying almost entirely within the span of the other active directions, where exact decorrelation leaves little of the target behind. This is the geometric mechanism described in the main text. 

\begin{table}[t]
\centering
\small
\begin{tabular}{lcccc}
\toprule
Target & Naive & Ginv & Gain & Wilcoxon $p$ \\
\midrule
Accuracy        & $-33.4\%$ & $-3.7\%$  & $+29.7$ & $0.000$ \\
Ecology         & $-10.0\%$ & $+0.6\%$  & $+10.6$ & $0.127$ \\
Efficiency      & $-15.7\%$ & $-6.7\%$  & $+9.0$  & $0.181$ \\
Equity\_Future  & $+2.6\%$  & $+7.1\%$  & $+4.6$  & $0.663$ \\
Equity\_Present & $+6.0\%$  & $+36.0\%$ & $+30.0$ & $0.001$ \\
Precaution      & $+11.6\%$ & $+11.4\%$ & $-0.1$  & $0.994$ \\
Resilience      & $+3.7\%$  & $+1.4\%$  & $-2.3$  & $0.768$ \\
TechOpt         & $+7.4\%$  & $+26.0\%$ & $+18.6$ & $0.064$ \\
\midrule
MACRO           & $-3.5\%$  & $+9.0\%$  & $+12.5$ & --- \\
\bottomrule
\end{tabular}
\caption{Per-target NSE at $K=8$ (climate). Naive steering drives several targets negative (Accuracy $-33.4\%$); the correction recovers them. Significance by Wilcoxon signed-rank.}
\label{tab:kscaling-k8}
\end{table}

\paragraph{Compression and conditioning.} The $K=12$ row is the uncompressed taxonomy, and it settles the second question. The macro gain survives it ($+6.0\%$ naive to $+17.8\%$ corrected), so compression is not what makes the correction work. What compression buys is conditioning. Where directions are nearly parallel, the closed-form inverse cancels large, nearly equal quantities against one another, and the coefficient norm grows with $\kappa(\mat{G})$; this is the effect the adaptive relaxation of Appendix~\ref{app:hbeta} is built to control, and at $\kappa(\mat{G}) = 329.7$ the twelve-dimension set sits well inside that regime. Grouping correlated values before steering is a choice about conditioning rather than a concession that the correction cannot handle correlation.

\paragraph{Takeaway.} The condition number of $\mat{G}$ rises steeply with $K$, confirming that joint multi-value steering becomes geometrically harder as more values are added. Against that trend the inverse-Gram correction holds a positive and significant macro gain at every $K$, at least as large as at $K=4$. The per-target results show the source of the gain: it is largest on the dimensions that naive steering drives most negative, which is what the geometric account predicts. The $K=12$ row answers the second question. The macro gain survives the raw taxonomy ($+6.0\%$ to $+17.8\%$), so consolidation is not what makes the correction work. What consolidation buys is conditioning: at $\kappa(\mat{G}) = 329.7$ the closed-form inverse must cancel large, nearly parallel components against one another, which inflates the coefficient norm (Appendix~\ref{app:hbeta}). Grouping correlated values before steering is a practical choice about conditioning rather than a concession that the correction cannot handle correlation.

\section{From-scratch replication on a new domain: medical ethics}
\label{app:medical}

To test whether the method generalizes beyond climate, we built a second value domain from scratch, in an area with an established and independently motivated set of values: medical ethics. We began from seven candidate values, the four Beauchamp--Childress principles (Autonomy, Beneficence/CareDuties, Non-maleficence, Justice/DistributiveEthics) together with Evidence, Public Health, and Precaution, and applied the same taxonomy-compression procedure used for climate (Appendix~\ref{app:taxonomy}), which merged them into four dimensions: Autonomy, CareDuties, DistributiveEthics, and Evidence. We extracted one contrastive steering direction per dimension and evaluated with the same judge protocol as the main experiment ($50$ questions, three judgments per comparison with randomized A/B position and majority vote). Judge annotations agreed with a held-out check at $96.6\%$.

\paragraph{A well-conditioned domain.} The medical-ethics directions are close to orthogonal: the Gram matrix has condition number $\kappa(\mat{G}) = 5.30$, well below the climate value of $24.4$ at the same $K=4$. This follows from how the principles were defined; the Beauchamp--Childress principles were formulated to be conceptually independent. It makes the domain a test in the opposite direction from $K$-scaling. Here the geometry is benign and naive steering performs reasonably on its own, so the question is whether the correction still adds anything.

\paragraph{Macro NSE.} Table~\ref{tab:medical} reports macro NSE. Naive steering already reaches $+2.5\%$, higher than in the more entangled climate setting, and the inverse-Gram correction reaches $+14.2\%$, a $+11.7$pp gain (Wilcoxon $p=0.0043$, paired bootstrap $p=0.0040$), larger than the $+8.1$pp climate gain at the same $K$. The correction improves dimension-specific control even when the directions are already fairly separable, so its usefulness does not depend on pathological geometry.

\begin{table}[t]
\centering
\small
\begin{tabular}{lcccc}
\toprule
Dimension & Naive & Ginv & Gain & Wilcoxon $p$ \\
\midrule
Evidence           & $-2.7\%$  & $+32.6\%$ & $+35.3$ & $0.0006$ \\
Autonomy           & $-8.6\%$  & $+4.5\%$  & $+13.2$ & $0.045$  \\
DistributiveEthics & $+11.0\%$ & $+14.0\%$ & $+3.0$  & $0.187$  \\
CareDuties         & $+10.3\%$ & $+5.7\%$  & $-4.6$  & $0.62$   \\
\midrule
MACRO              & $+2.5\%$  & $+14.2\%$ & $+11.7$ & $0.0043$ \\
\bottomrule
\end{tabular}
\caption{Per-dimension and macro NSE on the from-scratch Medical Ethics domain ($n=50$, $\kappa(\mat{G})=5.30$). The gain concentrates on Evidence, the dimension most orthogonal to the others; the three related normative dimensions gain less. Significance by Wilcoxon signed-rank.}
\label{tab:medical}
\end{table}

\paragraph{Per-dimension structure.} The gain is uneven across dimensions, in the way the geometry predicts. It is dominated by Evidence ($+35.3$pp), the dimension most orthogonal to the rest, since it concerns empirical grounding rather than a normative stance and its steering direction shares little with the others, so the correction can act on it with little interference. The three normative dimensions (Autonomy, CareDuties, DistributiveEthics) are conceptually related and more collinear, so the correction has less room to separate them and their individual gains are small, or for CareDuties slightly negative. This matches the per-target pattern under $K$-scaling (Appendix~\ref{app:kscaling}), where the correction's benefit is largest on structurally separable dimensions and smallest on dimensions that overlap heavily with the others. The same signature appears in an independently constructed domain with a different value set, which indicates that it reflects the method's geometry rather than anything specific to the climate taxonomy.

\paragraph{Takeaway.} A from-scratch replication in medical ethics, a domain with a benign near-orthogonal geometry and an externally motivated value set, yields a significant $+11.7$pp macro gain, larger than the climate gain at the same $K$. With the $K$-scaling results, this shows the correction generalizes along two independent axes: to more value dimensions, where geometry worsens, and to different domains, where geometry may instead be benign, with a per-dimension benefit that tracks orthogonality in both.
\section{Judge reliability and cross-family agreement}
\label{app:crossjudge}

Because our primary metric relies on an LLM judge, we assess its reliability in two ways: whether the method-level conclusions hold under judges from different model families, and how closely individual judgments agree across judges and across phrasings of the judge instruction. The two questions have different answers, and the distinction matters for interpreting the results.

\paragraph{Method-level agreement.} We re-judged the full nine-method matrix ($50$ questions, all four direction types under both correction strategies, plus the layer-separation baseline) with GPT-4o, and cross-checked the headline comparison with GPT-5-mini. Table~\ref{tab:crossjudge-methods} reports NSE under DeepSeek and GPT-4o for all nine methods. The two judges agree closely at the method level: the headline method, contrast with the inverse-Gram correction, scores $+15.25\%$ under DeepSeek and $+15.42\%$ under GPT-4o, and the ordering in which the correction improves over naive steering holds under both judges for every direction type except probe under GPT-4o. GPT-5-mini, on the contrast comparison over $100$ questions, gives a gain of $+7.42$pp (paired bootstrap $p=0.0016$), consistent in sign and magnitude with the main result.

\begin{table}[t]
\centering
\small
\begin{tabular}{lcc}
\toprule
Method & DeepSeek NSE & GPT-4o NSE \\
\midrule
probe\_naive    & $+1.50\%$  & $+8.08\%$ \\
probe\_ginv     & $+3.83\%$  & $+5.58\%$ \\
contrast\_naive & $+4.00\%$  & $+3.75\%$ \\
contrast\_ginv  & $+15.25\%$ & $+15.42\%$ \\
actadd\_naive   & $-1.67\%$  & $+3.67\%$ \\
actadd\_ginv    & $+4.00\%$  & $+5.83\%$ \\
repe\_naive     & $+5.33\%$  & $+5.42\%$ \\
repe\_ginv      & $+6.75\%$  & $+7.92\%$ \\
layer\_sep      & $+2.75\%$  & $+0.83\%$ \\
\bottomrule
\end{tabular}
\caption{Per-method NSE under two judges from different families (DeepSeek and GPT-4o), on the same $50$ questions. The correction improves over naive steering under both judges for contrast, ActAdd, and RepE; the one exception is probe under GPT-4o.}
\label{tab:crossjudge-methods}
\end{table}

\paragraph{Item-level agreement.} At the level of individual comparisons the two judges agree less closely. Over all $7{,}200$ comparison cells, the raw three-way agreement (steered / baseline / tie) is $59.2\%$ and Cohen's $\kappa = 0.393$. Excluding ties, binary agreement on the remaining cells is $73.0\%$. We report this openly: single-item value judgments are genuinely close calls, and moderate agreement is what one should expect from them, for automated judges.

\paragraph{Ordinal-weighted agreement.} Because the three labels are ordered (baseline, tie, steered), we also computed Krippendorff's $\alpha$ with ordinal weighting, which penalizes a baseline-versus-steered disagreement more than a tie-versus-steered one. The nominal $\alpha = 0.3814$ matches Cohen's $\kappa$ as expected (both measure two-rater agreement on the same $7{,}200$ units). The ordinal $\alpha = 0.3494$ is slightly lower rather than higher. A tie-boundary explanation would have raised it; the drop instead indicates that many disagreements are genuine opposite calls on the harder items. This is consistent with the reading that per-item value judgments are subjective, while leaving the aggregate conclusion intact.

\paragraph{Sensitivity to judge-prompt phrasing.} Item-level judgments are also sensitive to the phrasing of the judge instruction, in the same range as they are sensitive to judge family. We re-judged the headline pair head-to-head under three instruction phrasings: the paper's wording, verbatim (``Which response places MORE emphasis on [dimension]?''), an expression variant (``which response expresses this value more strongly?''), and a prioritization variant (``One of these two answers prioritizes [dimension] more than the other. Which one?''), each with its own answer-format line. For each of 50 (target dimension, question) items, the judge compares the corrected and the naive response directly and selects which places more emphasis on the item's target dimension (three DeepSeek calls per phrasing, A/B position independently re-randomized under each phrasing, majority vote), reusing the main-experiment generations ($\alpha = 26$, layer 14). The items are a stratified sample, balanced by construction: 20 favorable to each arm and 10 neutral under a proxy from the main-experiment outcomes. The head-to-head preference rate is therefore an instrument for measuring wording effects, not an estimate of either method's superiority, and it moves substantially with the wording: the corrected arm is preferred on 20, 24, and 33 of the 50 items (40\%, 48\%, 66\%) under the three phrasings, with no majority-vote ties, and item-level agreement between phrasings is 46.0--60.0\%, in the same range as the 59.2\% raw agreement between judge families above. This head-to-head also measures a different quantity from NSE: it compares the two arms on target emphasis alone, where Table~\ref{tab:main_3b} shows them nearly equal (target win rates 56.8\% against 55.2\%), while the correction's gain is concentrated in spillover suppression (42.9\% against 49.4\%), which no arm-versus-arm target comparison can see. Individual close-call judgments move with phrasing as they move with judge family, consistent with the moderate $\kappa$ above.

\paragraph{A local logprob judge.} The gain also survives a change of judging mechanism, from sampled text verdicts to a deterministic log-probability readout on a local open-weights model. Llama-3.1-8B-Instruct, run locally, judges each comparison once under the main judge instruction, taking the most probable of the three answer tokens (A, B, tie) as the verdict; this removes API dependence, sampling noise, and provider-side drift. Over the full contrast matrix (two methods, four target dimensions, 200 questions, all four judged dimensions; $6{,}400$ comparisons), contrast\_naive reaches NSE $+4.0\%$ (target $52.3\%$, spillover $48.3\%$) and contrast\_ginv $+13.5\%$ (target $58.4\%$, spillover $44.9\%$), a $\mat{G}^{-1}$ gain of $+9.5$pp ($95\%$ CI $[+5.3, +13.9]$), consistent with DeepSeek's $+8.1$pp on the same pair. Item-level agreement with the DeepSeek majority verdicts on the identical $6{,}400$ comparisons is $55.6\%$ raw with Cohen's $\kappa = 0.353$, the same moderate range as the API judge pair above: a fourth judge family and a different readout mechanism move individual close calls and leave the method-level conclusion intact.

\paragraph{Aggregation over close calls.} The item-level noise averages out at the method level. Individual comparisons are close calls on which two judges, two phrasings of the instruction, or two people often differ, but the aggregate NSE of a method is estimated over hundreds of such comparisons, and there the two judges agree closely (Table~\ref{tab:crossjudge-methods}). The conclusion that the inverse-Gram correction reduces net spillover rests on the aggregate, which is stable across judges, not on any single comparison.
\section{Robustness across layers}
\label{app:layer_sweep}

The main experiments intervene at one layer per model, chosen by probe F1 during setup with no reference to steering gain (layer~14 on Llama-3.2-3B-Instruct). A natural concern is that the correction's benefit could be specific to that layer. We repeated the contrast comparison at $L = 10$, $14$, and $18$ on the 3B model.

\paragraph{Protocol.} Directions are re-extracted at each layer from the same contrast data, so every layer has its own direction set and Gram matrix. Steering strength is held fixed at the paper's $\alpha = 26$ at all three layers; we deliberately do not re-calibrate $\alpha$ per layer, so that no per-layer tuning enters the comparison. Evaluation uses the first 50 canonical questions and the main judge protocol (DeepSeek, three judgments per comparison with randomized A/B position and majority vote). $L = 10$ and $L = 18$ are freshly generated and judged ($n = 50$); the $L = 14$ row sub-selects the same 50 questions from the main-experiment judgments, so its values coincide with the DeepSeek subset entries in Appendix~\ref{app:crossjudge}.

\paragraph{Results.} Table~\ref{tab:layer_sweep} reports the results. The inverse-Gram correction yields a positive gain at all three layers: $+3.00$pp at $L=10$, $+11.25$pp at $L=14$, and $+6.08$pp at $L=18$. The benefit is a property of the correction, and does not depend on the specific layer choice.

\paragraph{Interpretation.} The gain magnitude varies across layers largely because the naive baseline itself varies: at $L = 18$ naive steering already reaches $+8.00\%$, leaving less headroom, while at $L = 14$ it sits at $+4.00\%$. The condition number is also non-monotone in depth ($33.33$, $24.37$, $27.75$). We therefore claim that the advantage is robust across this layer range; we do not claim that the gain grows with depth or with $\kappa(\mat{G})$.

\begin{table}[t]
\centering
\small
\begin{tabular}{ccccc}
\toprule
Layer & $\kappa(\mat{G})$ & Naive & Ginv & Gain \\
\midrule
10 & 33.33 & $+5.50\%$ & $+8.50\%$  & $+3.00$pp  \\
14 & 24.37 & $+4.00\%$ & $+15.25\%$ & $+11.25$pp \\
18 & 27.75 & $+8.00\%$ & $+14.08\%$ & $+6.08$pp  \\
\bottomrule
\end{tabular}
\caption{Layer sweep on Llama-3.2-3B-Instruct: contrast directions, fixed $\alpha = 26$, first 50 canonical questions. Directions and the Gram matrix are re-extracted at each layer. The $L=14$ row sub-selects the main-experiment judgments on the same questions.}
\label{tab:layer_sweep}
\end{table}

\section{Response quality under steering}
\label{app:quality}

The correction amplifies steering coefficients ($\|\mat{G}^{-1}\vect{w}\| \geq \|\vect{w}\|$, growing with entanglement), so it is reasonable to ask whether decorrelation costs generation quality relative to naive steering, and what steering of either kind costs relative to the unsteered model. We measure both directly, in the worst case for the correction: contrast directions, the most entangled family ($\kappa = 24.4$), where coefficient amplification is largest.

\paragraph{Protocol.} We evaluate the first 50 canonical questions under three arms, reusing the main-experiment generations at $\alpha = 26$, layer~14: the unsteered baseline, contrast\_naive, and contrast\_ginv. Perplexity is the teacher-forced negative log-likelihood of the response string alone, with no chat template and truncation at 1024 tokens, scored by Llama-3.2-3B-Instruct with the steering hook removed. The same scorer is applied to all three arms, so the comparison is internally consistent, although self-scored perplexity is a relative measure rather than an absolute one. Fluency is judged by GPT-4o on a 1-to-5 scale with an instruction to rate writing quality only (grammatical correctness, logical flow, readability, absence of repetition or degenerate text) and to disregard content and topic emphasis; one judgment per response at temperature~0, one response per question per arm.

\paragraph{Results.} Median perplexity is 2.36 for naive steering and 2.55 for the corrected condition, a median difference of $+0.19$; mean perplexity is 2.27 for the unsteered baseline, 2.48 for naive, and 2.60 for corrected. Mean fluency is 4.16 unsteered, 3.98 naive, and 4.00 corrected.

\paragraph{Reading.} The corrected-minus-naive differences, $+0.19$ in median perplexity and $+0.02$ in mean fluency, show that decorrelation adds no measurable fluency cost and only a small perplexity increase over naive steering, even for the most entangled direction family. Steering of either kind has a bounded cost relative to the unsteered model: under 0.2 fluency points and between $+0.2$ and $+0.3$ in mean perplexity. The steered text is fluent rather than degenerate; the qualitative pairs in the main text show the effect of the correction on content emphasis with the writing quality intact.

\section{Adaptive regularization for collinear preferences}
\label{app:hbeta}

Eq.~\ref{eq:decorrelation} shows that $\mat{H} = \mat{G}^{-1}$ achieves exact projection matching for any $\vect{w}$. Appendix~\ref{app:mixed_w} shows the behavioral price when the request itself asks nearly collinear directions to move together: exact separation of directions at cosine $0.76$ requires large opposing coefficients, and the corrected condition underperforms naive steering there. This appendix defines the adaptive relaxation referenced in that discussion and characterizes the trade-off it navigates.

\paragraph{Definition.} For $\beta \geq 0$, let
\begin{align*}
\mat{H}_\beta = (\mat{G} + \beta \mat{I})^{-1}.
\end{align*}
At $\beta = 0$ this is the paper's correction. As $\beta$ grows, $\mat{H}_\beta \to \tfrac{1}{\beta}\mat{I}$, which is naive steering up to scale. In the eigenbasis $\mat{G} = \sum_i \lambda_i u_i u_i^\top$, we have $\mat{H}_\beta = \sum_i (\lambda_i + \beta)^{-1} u_i u_i^\top$: the regularization shrinks the correction most along the small-eigenvalue directions of $\mat{G}$, which are the ill-conditioned components responsible for coefficient amplification. $\mat{H}_\beta$ is closed-form and costs the same single $K \times K$ inverse as $\mat{G}^{-1}$.

\paragraph{Behavioral motivation.} Two independent measurements locate where $\beta = 0$ stops helping. In the mixed-$\vect{w}$ experiment (Table~\ref{tab:mixed_w}), the corrected-minus-naive advantage over the three two-dimensional configurations falls monotonically with the within-support cosine: $+10.50$pp at $\cos = 0.54$, $+4.50$pp at $0.55$, $-1.75$pp at $0.76$. A targeted two-pair replication with $0.5/0.5$ weights ($n = 50$ per configuration, $\alpha = 26$, main judge protocol) shows the same reversal: at $\cos(d_{\mathrm{Acc}}, d_{\mathrm{TE}}) = 0.62$ the correction wins ($+5.5\%$ against $-0.5\%$ NSE), while at $\cos(d_{\mathrm{SJ}}, d_{\mathrm{FE}}) = 0.87$ naive steering wins ($+15.5\%$ against $+3.5\%$). In our data the correction is positive at within-support cosines up to $0.62$ and negative at $0.76$ and above, on this model and direction family.

\paragraph{Geometric trade-off.} Table~\ref{tab:hbeta} reports, for the paper's 3B contrast Gram matrix (Table~\ref{tab:gram_3b}), the coefficient vector $\vect{\alpha}_\beta = \mat{H}_\beta \vect{w}$ and the achieved projections $\mat{G}\vect{\alpha}_\beta$ for two requests. At $\beta = 0$ the projections equal $\vect{w}$ exactly, at the price of large coefficients: $\|\vect{\alpha}_\beta\| = 3.13$ for the TE{+}FE request and $5.86$ for one-hot SocialJustice, against $\|\vect{w}\| = 0.71$ and $1$. Increasing $\beta$ shrinks $\|\vect{\alpha}_\beta\|$ and the conditioning, and off-support projection returns in order of cosine with the active set: for one-hot SocialJustice, the leakage at $\beta > 0$ is largest on FutureEthics ($\cos = 0.87$), then TechEcon ($0.77$), then Accuracy ($0.54$). The trade is not free in the other direction either: $\|\vect{\alpha}_\beta\|$ falls faster than the leakage returns, so heavy regularization shrinks the overall steering signal more than it buys back in separation. $\beta$ therefore has to be matched to the request: near zero when the active set is well separated, positive when the request is internally collinear. Because the within-support cosines are entries of $\mat{G}$, this choice is computable at request time at no additional cost.

\paragraph{Status.} $\mat{H}_\beta$ is defined in closed form, the geometry above characterizes the trade-off it navigates, and the behavioral evidence locates where relaxation is needed. A behavioral evaluation of request-adaptive $\beta$ schedules is left to future work; we do not claim here that $\mat{H}_\beta$ recovers the collinear cases.

\begin{table}[t]
\centering
\small
\begin{tabular}{lcc|cccc}
\toprule
$\beta$ & $\kappa(\mat{G}+\beta\mat{I})$ & $\|\vect{\alpha}_\beta\|$ & Acc & TE & SJ & FE \\
\midrule
\multicolumn{7}{c}{$\vect{w} = (0, 0.5, 0, 0.5)$ \quad (TE{+}FE)} \\
\midrule
0    & 24.37 & 3.13 & 0.000 & 0.500 & 0.000 & 0.500 \\
0.1  & 14.02 & 1.82 & 0.038 & 0.423 & 0.119 & 0.394 \\
0.5  & 5.70  & 0.71 & 0.090 & 0.306 & 0.191 & 0.288 \\
2    & 2.38  & 0.24 & 0.095 & 0.187 & 0.155 & 0.183 \\
\midrule
\multicolumn{7}{c}{$\vect{w} = e_{\mathrm{SJ}}$ \quad (one-hot SocialJustice)} \\
\midrule
0    & 24.37 & 5.86 & 0.000 & 0.000 & 1.000 & 0.000 \\
0.1  & 14.02 & 3.33 & 0.013 & 0.074 & 0.720 & 0.165 \\
0.5  & 5.70  & 1.26 & 0.056 & 0.147 & 0.435 & 0.235 \\
2    & 2.38  & 0.40 & 0.077 & 0.136 & 0.234 & 0.175 \\
\bottomrule
\end{tabular}
\caption{Regularization path of $\mat{H}_\beta = (\mat{G} + \beta\mat{I})^{-1}$ on the 3B contrast Gram matrix. Projections are $\mat{G}\vect{\alpha}_\beta$, the achieved emphasis along each direction; $\kappa(\mat{G} + \beta\mat{I})$ depends only on $\beta$. At $\beta = 0$ the projections equal the request exactly; larger $\beta$ shrinks the coefficient norm and reintroduces off-support projection in order of cosine with the active set.}
\label{tab:hbeta}
\end{table}

\section{Instruction prompting as baseline and complement}
\label{app:prompting}

Instructing the model to emphasize a value is the zero-training alternative to activation steering, and it is a strong one: on single-value NSE it is statistically indistinguishable from corrected steering. It is also a complement rather than a substitute. Its strength depends heavily on wording, no weights appear in its instructions, and stacking it with corrected steering at a re-calibrated strength exceeds either alone at matched response quality. This appendix reports the comparison, the phrasing sensitivity, and the stack.

\paragraph{Protocol.} We test three phrasings of the single-value instruction, injected as the system prompt with the question left unmodified. P1: ``Please emphasize \{value\} in your response.'' P2: ``You should emphasize \{value\} in your response. Focus on this dimension when answering: \{description\}.'' P3 assigns a persona, e.g. for SocialJustice ``You are an environmental justice advocate who centers equity, vulnerable communities, and fair distribution of burdens.'', with analogous personas for the other dimensions (a scientist prioritizing empirical evidence, a technology optimist and economist, an intergenerational ethicist). The \{value\} strings are ``scientific accuracy and evidence-based reasoning'' (Accuracy), ``technological solutions and economic efficiency'' (TechEcon), ``social equity and fairness'' (SocialJustice), and ``long-term sustainability and intergenerational ethics'' (FutureEthics); the mixed-$\vect{w}$ prompting condition of Appendix~\ref{app:mixed_w} uses the same strings. Prompted responses are generated by sampling (temperature 0.7, top-$p$ 0.9, up to 200 new tokens); the corrected-steering arm reuses the main-experiment generations. Judging is identical to the main protocol: DeepSeek, three judge calls per comparison with randomized A/B position and majority vote, each arm judged against the unsteered baseline.

\paragraph{Single-value comparison.} Table~\ref{tab:prompting} reports per-dimension and macro NSE for prompting (P1) against contrast\_ginv on the full 200 canonical questions. Prompting is nominally higher on the macro ($+16.90\%$ against $+13.96\%$), and the difference is not statistically significant (paired bootstrap $p = 0.231$; Wilcoxon signed-rank on the same per-question pairing $p = 0.23$). Per dimension, the only significant difference favors prompting, on TechEcon ($p = 0.014$); the largest nominal advantage in the other direction is corrected steering on SocialJustice ($p = 0.163$). We claim single-value superiority for neither method.

\begin{table}[t]
\centering
\small
\begin{tabular}{lccc}
\toprule
Target dimension & contrast\_ginv & Prompting (P1) & $p$ \\
\midrule
Accuracy      & $+14.42\%$ & $+14.92\%$ & $0.924$ \\
TechEcon      & $+14.33\%$ & $+25.58\%$ & $0.014$ \\
SocialJustice & $+22.50\%$ & $+15.58\%$ & $0.163$ \\
FutureEthics  & $+4.58\%$  & $+11.50\%$ & $0.145$ \\
\midrule
MACRO         & $+13.96\%$ & $+16.90\%$ & $0.231$ \\
\bottomrule
\end{tabular}
\caption{Per-dimension and macro NSE for instruction prompting (P1) against corrected steering (contrast\_ginv) on the full 200 canonical questions (Llama-3.2-3B-Instruct, main judge protocol). $p$ is a paired bootstrap on the per-question difference; the macro difference is also not significant under a Wilcoxon signed-rank on the same pairing ($p = 0.23$).}
\label{tab:prompting}
\end{table}

\paragraph{Sensitivity to phrasing.} Prompting strength is a property of the wording. Across the three phrasings of the same request, on the same 50-question subset, macro NSE moves from $+12.58\%$ (P1) to $+21.50\%$ (P2) to $+29.17\%$ (P3), with corrected steering at $+15.25\%$ as the reference (its main-experiment judgments restricted to this subset); on SocialJustice alone the spread runs from $+8.00\%$ to $+42.67\%$. Decomposing NSE on this subset, corrected steering attains its effect at the lowest spillover win rate of the four arms ($42.50\%$, against $44.33$--$48.92\%$ for the three phrasings): the instructions lift non-target emphasis along with the target. Prompting exposes no explicit knob on this strength; on the steering side, strength is the scalar $\alpha$ (Appendix~\ref{app:alpha_sweep}) and allocation is $\vect{w}$, both independent of wording.

\paragraph{Stacking.} The two mechanisms compose, and the comparison is between operating points, each configuration at its calibrated strength. Table~\ref{tab:stacking} reports the ladder on the first 100 canonical questions. At the main experiment's $\alpha = 26$, adding the P1 instruction on top of corrected steering reaches the highest macro NSE ($+37.25\%$), exceeding corrected steering alone on all four dimensions (significant on three: bootstrap $p = 0.006$, $p < 0.001$, $p < 0.001$; FutureEthics $p = 0.12$) and prompting alone on all four (significant on two), but pays a perplexity cost (median 2.79). Because the instruction already supplies part of the push, halving the steering strength to $\alpha = 13$ retains $+27.71\%$ at quality parity with corrected steering at its own operating point: median perplexity 2.49 for the stack at $\alpha = 13$ against 2.48 for corrected steering at $\alpha = 26$, fluency 3.99 against 3.97. On the same questions and at the same response quality, the stack roughly doubles the macro NSE of corrected steering alone.

\begin{table}[t]
\centering
\small
\begin{tabular}{lccc}
\toprule
Configuration & Macro NSE & Median PPL & Fluency \\
\midrule
Unsteered baseline & --- & 2.19 & 4.15 \\
Prompting (P1) & $+17.08\%$ & 2.32 & 4.03 \\
contrast\_ginv ($\alpha = 26$) & $+12.88\%$ & 2.48 & 3.97 \\
Stack ($\alpha = 26$) & $+37.25\%$ & 2.79 & 4.01 \\
Stack ($\alpha = 13$) & $+27.71\%$ & 2.49 & 3.99 \\
\bottomrule
\end{tabular}
\caption{Operating-point ladder on the first 100 canonical questions (Llama-3.2-3B-Instruct, contrast directions, layer 14). Macro NSE from the powered stacking run ($n = 100$ questions per target dimension); the prompting and corrected-steering reference rows reuse the corresponding judgments restricted to this subset, hence the expected subset variation from Table~\ref{tab:prompting}. Median perplexity and mean fluency (1--5) are computed on the same subset (100 questions; 400 generations per four-dimension arm); both metrics follow the protocol of Appendix~\ref{app:quality}. NSE is measured against the unsteered baseline, so the baseline row has none.}
\label{tab:stacking}
\end{table}

\paragraph{Granularity of control.} The instruction templates make the structural difference visible: no weights appear in any of them. A prompt names the values to emphasize; a preference ratio such as $0.7/0.3$ never enters the text, and the mixed-$\vect{w}$ instruction of Appendix~\ref{app:mixed_w} likewise names only the active set. The continuous $\vect{w}$ buys programmatic composability and reproducibility instead: it enters the intervention numerically through $\vect{\alpha} = \mat{G}^{-1}\vect{w}$, composes linearly across dimensions, stacks with prompting, and is invariant to wording, whereas the results above show prompting strength varying by a factor of two across natural phrasings of the same request. We make no claim that nearby weight vectors produce reliably distinguishable individual outputs; at the pattern level, the fidelity analysis of Appendix~\ref{app:mixed_w} shows that the achieved emphasis profile tracks the requested $\vect{w}$.

\section{Human validation of the pairwise emphasis judgments}
\label{app:human}
 
Every NSE number in this paper rests on pairwise emphasis judgments made by an LLM judge. This appendix validates those judgments against human annotators performing the same unit task under the same instruction: given a question, two responses in randomized left-right order, and one named value dimension, select the response that places more emphasis on that dimension, or call a tie. The instruction wording matches the judge prompt (Appendix~\ref{app:eval}), and annotation is blind, with nothing marking which response is steered.
 
\paragraph{Sample.} The sample contains 240 comparisons drawn from the primary 3B experiment, reusing the main-experiment generations (contrast directions, $\alpha = 26$, layer~14). Each item pairs a steered response with the unsteered baseline response to the same question and names one judged dimension; 115 items come from contrast\_ginv and 125 from contrast\_naive. Sampling is stratified into 96 target items, where the judged dimension is the steering target, and 144 spillover items, where it is one of the three others. All four judged dimensions are covered (Accuracy 65, SocialJustice 64, TechEcon 58, FutureEthics 53), and the steered response appears on the left in 127 of the 240 items.
 
\paragraph{Annotators.} Five annotators labelled the sample; three covered all 240 items and two covered initial contiguous blocks. One annotator whose labels correlated near zero with both the other annotators and the judge was excluded as an outlier, leaving four. Inter-annotator agreement is computed on the three annotators who labelled every item; the steered-selection rate and human-judge agreement use the majority label over all four retained annotators wherever available.
 
\paragraph{Measures.} We report three quantities: agreement between the human labels and the judge's verdicts on the same items, the direct validation target; inter-annotator agreement, on the same chance-corrected scale as the cross-judge comparison (Appendix~\ref{app:crossjudge}); and the human counterpart of the steering readout, the rate at which annotators select the steered response over the baseline, split by item type and method. Effective steering appears as a higher steered-selection rate on target items, and spillover suppression appears as a lower steered-selection rate on spillover items for contrast\_ginv than for contrast\_naive.
 
\paragraph{Results.} Table~\ref{tab:human} reports the results, and they support the judge on both axes. Humans and the judge agree at $\kappa = 0.51$ on the identical comparisons, above the $\kappa = 0.39$ the three judge families reach among themselves (Appendix~\ref{app:crossjudge}): the automated verdicts track human judgment at least as closely as human-calibrated judges track each other. Agreement among the three full-coverage annotators is moderate, $\kappa = 0.41$, in the same range and consistent with the subjectivity of a single emphasis comparison. The steering readout carries through to human labels: annotators select the steered response more than twice as often for contrast\_ginv as for contrast\_naive on target items ($0.33$ against $0.15$), and on spillover items the steered-selection rate stays low for both ($0.14$ against $0.12$), the same pattern the automated NSE reports. Human evaluation confirms both that the judge is a faithful stand-in and that the correction behaves as claimed.
 
\begin{table}[t]
\centering
\small
\begin{tabular}{lcc}
\toprule
Steered-selection rate & contrast\_naive & contrast\_ginv \\
\midrule
Target items      & $0.149$ & $0.327$ \\
Spillover items   & $0.115$ & $0.136$ \\
\midrule
Agreement & Raw & $\kappa$ \\
\midrule
Human--judge ($n = 240$)                    & $0.704$ & $0.508$ \\
Inter-annotator (full-coverage)$^{\dagger}$ & $0.643$ & $0.405$ \\
\bottomrule
\end{tabular}
\caption{Human validation of the pairwise emphasis judgments. Top: rate at which annotators select the steered response over the unsteered baseline (majority vote; ties counted in the denominator), by item type and method. Bottom: agreement of the human labels with the LLM judge, and among annotators. Human--judge agreement ($\kappa = 0.51$) exceeds the cross-judge level ($\kappa = 0.39$), and inter-annotator agreement is moderate. One outlier annotator, whose labels correlated near zero with both the other annotators and the judge, was excluded. $^{\dagger}$Inter-annotator agreement (Fleiss $\kappa$, three-way) is over the three annotators who labelled all 240 items; the human--judge and steered-selection figures additionally use a fourth annotator who labelled 48 items.}
\label{tab:human}
\end{table}

\end{document}